\documentclass[journal]{IEEEtran}

\pdfoutput = 1

\usepackage{times}
\usepackage{soul}
\usepackage{url}
\usepackage{hyperref}
\usepackage[utf8]{inputenc}
\usepackage[small]{caption}
\usepackage{amsfonts,amssymb}
\usepackage{graphicx}
\usepackage{amsmath}
\usepackage{booktabs}
\usepackage{algorithm}
\usepackage{algorithmic}
\usepackage{multirow}
\usepackage{xspace}
\usepackage{subfigure}
\usepackage{footnote}
\usepackage{color}
\usepackage{wasysym}

\let\oldhat\hat

\renewcommand{\hat}[1]{\oldhat{\mathbf{#1}}}
\renewcommand{\matrix}[1]{\mathbf{#1}}

\usepackage{epstopdf}
\usepackage{threeparttable}
\usepackage{dcolumn}
\newcolumntype{d}[1]{D{.}{.}{#1}}

\newcommand{\eg}{\emph{e.g.,}\xspace}

\newcommand{\ie}{\emph{i.e.,}\xspace}
\newcommand{\etc}{\emph{etc.}\xspace}
\newcommand{\etal}{\emph{et al.}\xspace}
\newcommand{\aka}{\emph{a.k.a.,}\xspace}

\newcommand{\eat}[1]{}
\newcommand{\paratitle}[1]{\vspace{1ex}\noindent \textbf{#1}}
\begin{document}

\title{A Survey on Recent Advances in Sequence Labeling from Deep Learning Models}

\author{Zhiyong He, Zanbo Wang, Wei Wei\IEEEauthorrefmark{1}, Shanshan Feng, Xianling Mao, and Sheng Jiang

\thanks{This work was supported in part by the National Natural Science Foundation of China under Grant No. 61602197 and Grant No. 61772076, Grant No. 61972448, Grant No. L1924068 and in part by Equipment Pre-Research Fund for The 13th Five-year Plan under Grant No. 41412050801.}
\thanks{E-mail addresses: weiw@hust.edu.cn (W. Wei)}
\thanks{Z. He is with the School of Electronic Engineering, Naval University of Engineering.}
\thanks{Z. Wang, W. Wei and S. Jiang are with the School of Computer Science and Technology, Huazhong University of Science and Technology.}
\thanks{S. Feng is with the Inception Institute of Artificial Intelligence Abu Dhabi, UAE.}
\thanks{X. Mao is with the School of Computer, Beijing Institute of Technology.}

}

%
%


\maketitle

\begin{abstract}
Sequence labeling (SL) is a fundamental research problem encompassing a variety of tasks,
\eg part-of-speech~(POS) tagging, named entity recognition~(NER), text chunking \etc
%
Though prevalent and effective in many downstream applications (\eg information retrieval, question answering and knowledge graph embedding),
conventional sequence labeling approaches heavily rely on hand-crafted or language-specific features.
%
Recently, deep learning has been employed for sequence labeling task due to its powerful capability in
automatically learning complex features of instances and effectively yielding the stat-of-the-art performances.
In this paper, we aim to present a comprehensive review of existing deep learning-based sequence labeling models,
which consists of  three related tasks, \eg  part-of-speech tagging, named entity recognition and text chunking.
%
Then, we systematically present the existing approaches base on a scientific taxonomy,
as well as the widely-used experimental datasets and popularly-adopted evaluation metrics in SL domain.
Furthermore, we also present an in-depth analysis of different SL models on the factors that may affect the performance,
and the future directions in SL domain.
\end{abstract}

\begin{IEEEkeywords}
  	Sequence labeling, deep learning, natural language processing.
\end{IEEEkeywords}

\IEEEpeerreviewmaketitle

\section{Introduction}


Sequence labeling is a type of pattern recognition task in the important branch of
natural language processing~(NLP).
From the perspective of linguistics,
the smallest meaningful unit in a language is typically regarded as morpheme,
and each sentence can thus be viewed as a sequence composed of morphemes.
Accordingly, the sequence labeling problem in NLP domain can be
formulate it as a task that aims at assigning labels to a category of morphemes
that generally have similar roles within the grammatical structure of sentences
and have similar grammatical properties,
%
and the meanings of the assigned labels usually depend on the types of specific tasks,
examples of classical tasks include
part-of-speech~(POS) tagging~\cite{Ma2016End}, named entity recognition~(NER)~\cite{Lample2016Neural}, text chunking~\cite{Liu2017Empower} and \etc,
which play a pivotal role in natural language understanding and
can benefit a variety of downstream applications such as
syntactic parsing~\cite{Nivre2004Deterministic}, relation extraction~\cite{Liu2017Heterogeneous} and entity coreference resolution~\cite{Ng2010Supervised} and \etc,
and hence has quickly gained massive attention.

%

Generally, conventional sequence labeling approaches are usually on the basis of classical machine learning technologies,
\eg Hidden Markov Models (HMM)~\cite{baum1966statistical} and Conditional Random Fields (CRFs)~\cite{lafferty2001conditional},
which often heavily rely on hand-crafted features (\eg whether a word is capitalized) or language-specific resources (\eg gazetteers).
Despite superior performance achieved, the requirement of the considerable amount of domain knowledge and efforts on feature engineering
make them extremely difficult to extend to new areas.
%
Over the past decade, the great success has been achieved by deep learning (DL) due to its powerful capability in automatically learning complex features of data.
Hence, there already exist many efforts dedicated to research on how to exploit the representation learning capability of deep neural network
for enhancing sequence labeling tasks,
and many of these methods have successively advanced the state-of-the-art performances~\cite{bohnet2018morphosyntactic,akbik2018contextual,devlin2018bert}.
This trend motivates us to conduct a comprehensive survey to summarize the
current status of deep learning techniques in the filed of sequence labeling.
By comparing the choices of different deep learning architectures, we aim to identify the impacts on the model performance, making it
convenient for subsequent researchers to better understand the advantages/disadvantages of such models.

\paratitle{Differences with former surveys}.
In literature, there have been numerous attempts to improve the performance of sequence labeling tasks using deep learning models.
However,
to the best of our knowledge,
there are nearly none comprehensive surveys that provide an in-depth summary of existing neural network based methods or developments on part of this topic so far.
%
Actually, in the past few years, several surveys have been presented on traditional approaches for sequence labeling.
For example, Nguyen \etal~\cite{nguyen2007comparisons} propose a systematic survey on machine learning based sequence labeling problems.
%
Nadeau \etal~\cite{nadeau2007survey} survey on the problem of \emph{named entity recognition} and present
an overview of the trend from hand-crafted rule-based algorithms to machine learning techniques.
Kumar and Josan~\cite{kumar2010part} conduct a short review on
\emph{part-of-speech tagging} for Indian language.
In summary, the most reviews for sequence labeling mainly cover papers on traditional machine learning methods,
rather than
the recent applied techniques of deep learning~(DL).
%
%
%
Recently, two work~\cite{yadav2018survey,li2018survey}
present a good literature survey of the deep learning based models for named entity recognition~(NER) problem,
however it is solely a sub-task for sequence labeling.
%
To the best of our knowledge, there has been so far no survey that can provide an exhaustive summary of recent research on DL-based sequence labeling methods.
Given the increasing popularity of deep learning models in sequence labeling, a systematic survey will be of high academic and practical significance. We summarize and analyze these related works and over $100$ studies are covered in this survey.

\paratitle{Contributions of this survey}.~The goal of this survey is to thoroughly review the recent applied techniques of
deep learning in the filed of sequence labeling~(SL),
and provides a panoramic view
to enlighten and guide the researchers and practitioners in SL research community
for quickly understanding and stepping into this area.
Specifically, we present a comprehensive survey on deep learning-based SL techniques
to systematically summarize the state-of-the-arts with a scientific taxonomy
along three axes, \ie
\emph{embedding module}, \emph{context encoder module}, and \emph{inference module}.
In addition, we also present an overview on the experiment settings (\ie \emph{dataset} or \emph{evaluation metric}) for commonly studied tasks in sequence labeling domain.
Besides, we have discussed and compared the results give by the most representative models for analyzing the effects
of different factors and architectures.
%
Finally, we present readers with the challenges and open issues faced by current DL-based sequence labeling methods and outline future directions in this area.

\paratitle{Roadmap}.~The remaining of this paper is organized as follows:
Section \ref{background} introduces background of sequence labeling, consisting of several related tasks and traditional machine learning approaches.
Section \ref{main} presents deep learning models for sequence labeling based on our proposed taxonomy.
Section \ref{datasets} summarizes the experimental settings (\ie dataset and evaluation metric) for related tasks.
%
Section \ref{results} lists the results of different methods,
followed by the discussion of the promising future directions.
Finally, Section \ref{conclusion} concludes this survey.


\section{Background}
\label{background}
%
In this section, we first give an introduction of three widely-studied classical \emph{sequence labeling} tasks,
\ie
\emph{part-of-speech~(POS)~tagging}, \emph{named entity recognition}~(NER) and \emph{text chunking}.
%
Then, we briefly introduce the traditional machine learning based techniques in sequence labeling domain.


\subsection{Classical Sequence Labeling Task}

\subsubsection{Part-of-speech Tagging (POS)}
POS receives a high degree of acceptance from both academia and industry,
which is a standard sequence labeling task
that aims at assigning a correct part-of-speech tag to each lexical item (\aka word)
such as noun (NN), verb (VB) and adjective (JJ).
In general, part-of-speech (POS) tagging can also be viewed as a subclass division of all words in a language,
which is thus also called a word class.
The tagging system of part-of-speech tags is not usually uniform under different data set,
\eg PTB (Penn Treebank)~\cite{marcus1993building},
%
%
which includes $45$ different types of POS tags
for word classification,
such as for sentence ``Mr. Jones is editor of the Journal",
it will be labeled with a sequence like "NNP NNP VBZ NN IN DT NN".

In fact, Part-of-speech can be regarded as a coarse-grained word cluster task,
the goal of which is to label the form and syntactic information of words in a sentence,
which is benefit for alleviating the sparseness of word-level features,
and servers as an important pre-processing step in natural language processing domain for various subsequent tasks
like semantic role labeling or syntax analysis.

\subsubsection{Named Entity Recognition~(NER)}
Name entity recognition (NER, \aka named entity identification or entity chunking), is a well-known classical sequence labeling task,
the goal of which is to identify named entities from text belonging to
pre-defined categories, which generally consists of three major categories (\ie entity, time, and numeric)
and seven sub-categories (\ie person name, organization, location, time, date, currency, and percentage).
%
Particularly, in this paper we mainly focus on the NER problem in English language,
and a widely-adopted English taxonomy is CoNLL2003 NER corpus~\cite{sang2003introduction},
which is collected from Reuters News Corpus that includes four different types of named entities, \ie  person (PER), location (LOC), organization (ORG) and proper nouns (MISC).

Generally, the label of a word in NER is composed of two parts, \ie ``X-Y'',
where ``X'' indicates the position of the labeled word and ``Y''
refers to the corresponding category within a pre-defined taxonomy.
%
In particular, it may be labeled with a special label (\eg ``none''), if a word cannot be classified into any pre-defined category.
%
Generally, the widely-adopted tagging scheme in the industry is BIOES system, that is, the word labeled ``B'' (Begin), ``I'' (Inside) and ``E'' (End) means that it is the first, middle or last word of a named entity phrase, respectively. The word labeled ``0-'' (Outside) means it does not belong to any named entity phrase and ``S-'' (Single) indicates it is the only word that represent an entity.

Named entity recognition is a very important task in natural language processing and is a basic technology for many high-level applications, such as search engine, question and answer systems, recommendation systems, translation systems, \etc Without loss of generality, we take machine translation as example to illustrate the importance of NER for various downstream tasks. In the process of translation, if the text contains named entity with a specific meaning, the translation system usually tends to translate multiple words that make up the named entity separately, resulting in blunt or even erroneous translation results. But if the named entity is identified first, the translation algorithm will have a better understanding of the word order and semantics of the text thus can output a better translation.

\subsubsection{Text Chunking}
The goal of the text chunking task is to divide text into syntactically related non-overlapping groups of words, \ie phrase, such as noun phrase, verb phrase, \etc The task can be essentially regarded as a sequence labeling problem that assign specific labels to words in sentences. Similar with NER, it can also adopt the BIOES tagging system. For example, the sentence ``The little dog barked at the cat.'' can be divided into the following phrases: ``(The little dog) (barked at) (the cat)''. Therefore, with the BIOES tagging system, the label sequence corresponding to this sentence is ``B-NP I-NP E-NP B-VP E-VP B-NP E-NP'', which means that ``The little dog'' and ``the cat'' are noun phrases and ``barted at'' is a verb phrase.

\subsubsection{Others}
There have been many explorations into applying the sequence labeling framework to address other problems such as dependency parsing~\cite{strzyz2019viable,li2018seq2seq}, semantic role labeling~\cite{park2019selectively,tan2018deep}, answer selection~\cite{zhou2015answer,li2016dataset}, text error detection~\cite{rei2016compositional,rei2016attending}, document summarization~\cite{nallapati2017summarunner}, constituent parsing~\cite{G2018Constituent}, sub-event detection~\cite{bekoulis2019sub}, emotion detection in dialogues~\cite{saxena2018emotionx} and complex word identification~\cite{gooding2019complex}.

\subsection{Traditional Machine Learning Based Approaches}
The traditional statistical machine learning techniques are the primary method for early sequence labeling problems. Based on the carefully designed features to represent each training data, the machine learning algorithms are utilized to train the model from example inputs and their expected outputs, learning to make predictions for unseen samples. Common statistical machine learning techniques include Hidden Markov Models (HMM)~\cite{eddy1996hidden}, Support Vector Machines (SVM)~\cite{hearst1998support}, Maximum Entropy Models~\cite{kapur1989maximum} and Conditional Random Fields (CRF)~\cite{lafferty2001conditional}.
HMM is a statistical model used to describe a Markov process with implicit unknown states. Bikel et al.~\cite{bikel1999algorithm} propose the first HMM-based model for NER system, named IdentiFinder. This model is extended by Zhou and Su~\cite{zhou2002named} and achieves better performance by assumimg mutual information independence rather than conditional probability independence of HMM.

SVM, which is alleged large margin classifier, is well-known for the good generalization capabilities and has been successfully applied to many pattern recognition problems. In the field of sequence labeling, Kudoh and Matsumoto~\cite{kudoh2000use} first propose to apply SVM classifier to the phrase chunking task and achieve the best performance at the time. Several subsequent studies using SVM for NER tasks are successively proposed~\cite{isozaki2002efficient,li2004svm}.

Ratnaparkhi~\cite{ratnaparkhi1996maximum} proposes the first maximum entropy model for part-of-speech tagging, and achieves great results. Some work for NER also adopt the maximum entry model~\cite{chieu2002named,bender2003maximum}. The maximum entropy markov model is further proposed~\cite{mccallum2000maximum}, which obtains a certain degree of improvement compared with the original maximum entropy model. Lafferty et al.~\cite{lafferty2001conditional} point out that utilizing the maximum entropy model for sequence labeling may suffer from a label bias problem. The proposed CRF model has achieved significant improvement in part-of-speech tagging and named entity recognition tasks and has gradually become the mainstream method of sequence labeling tasks~\cite{mccallum2003early,krishnan2006effective}.

\section{Deep Learning Based Models}
\label{main}

\begin{table*}[!t]
	\centering
     
	\caption{An overview of the deep learning based models for sequence lableing (LM: language model, pre LM emb: pretrained language model embedding, gaz: gazetteer, cap: capitalization, InNet: a
funnel-shaped wide CNN architecture~\cite{xin2018learning}, AE: autoencoder, MO-BiLSTM: multi-order Bi-LSTM~\cite{Zhang2017Does}, INN: implicitly-defined neural network~\cite{kazi2017implicitly}, EL-CRF: embedded-state latent CRF~\cite{thai2018embedded}, SRL: semantic role labeling), SA: self-attention}
   \setlength{\tabcolsep}{1pt}{
		\begin{tabular}{|c|c|c|c|c|c|c|}
			\hline
{\multirow{2}{*}{\bf Ref}}&\multicolumn{3}{c|}{\bf Embedding Module}
&{\multirow{2}{*}{\bf Context Encoder}}
&{\multirow{2}{*}{\bf Inference Module}}
&{\multirow{2}{*}{\bf Tasks}}
\\
\cline{2-4}&{\bf external input} &{\bf word embedding}&{\bf character-level}
& ~ & ~ & ~ \\

\hline
~\cite{Ma2016End} & $\backslash$ & Glove & CNN & Bi-LSTM & CRF & POS, NER \\
\hline
~\cite{bohnet2018morphosyntactic} & $\backslash$ & Word2vec & Bi-LSTM & Bi-LSTM & Softmax & POS \\
\hline
~\cite{Yasunaga2018Robust} & $\backslash$ & Glove & Bi-LSTM & Bi-LSTM & CRF & POS \\
\hline
~\cite{Liu2017Empower} & $\backslash$ & Glove & Bi-LSTM+LM & Bi-LSTM & CRF & POS, NER, chunking \\
\hline
~\cite{plank2016multilingual} & $\backslash$ & Polyglot & Bi-LSTM & Bi-LSTM & CRF & POS \\
\hline
~\cite{Rei2017Semi} & $\backslash$ & Word2vec & Bi-LSTM & Bi-LSTM+LM & CRF & POS, NER, chunking \\
\hline
~\cite{Peters2017Semi} & $\backslash$ & Senna & CNN & Bi-LSTM+ pre
LM & CRF & NER, chunking \\
\hline
~\cite{akbik2018contextual} & Pre LM emb & Glove & Bi-LSTM & Bi-LSTM & CRF & POS, NER, chunking \\
\hline
~\cite{zhang2018learning} & $\backslash$ & - & Bi-LSTM & Bi-LSTM & LSTM+Softmax & POS, NER \\
\hline
~\cite{ye2018hybrid} & $\backslash$ & Glove & Bi-LSTM+LM & Bi-LSTM & CRF+Semi-CRF & NER \\
\hline
~\cite{Zhang2017Does} & Spelling, gaz & Senna & $\backslash$ & MO-BiLSTM & Softmax & NER, chunking \\
\hline
~\cite{gregoric2018named} & $\backslash$ & Word2vec & Bi-LSTM & Parallel Bi-LSTM & Softmax & NER \\
\hline
~\cite{Yang2016Transfer} & $\backslash$ & Senna, Glove & Bi-GRU & Bi-GRU & CRF & POS, NER, chunking\\
\hline
~\cite{ling2015finding} & $\backslash$ & Trained on wikipedia & Bi-LSTM & Bi-LSTM & Softmax & POS \\
\hline
~\cite{chiu2016named} & Cap, lexicon  & Senna & CNN & Bi-LSTM & CRF & NER \\
\hline
~\cite{rei2016attending} & $\backslash$ & Word2vec & Bi-LSTM & Bi-LSTM & CRF & POS, NER, chunking \\
\hline
~\cite{xin2018learning} & $\backslash$ & Glove & InNet & Bi-LSTM & CRF & POS, NER, chunking \\
\hline
~\cite{kazi2017implicitly} & Spelling, gaz & Senna & $\backslash$ & INN & Softmax & POS\\
\hline
~\cite{thai2018embedded} & $\backslash$ & Glove & $\backslash$ & Bi-LSTM & EL-CRF & Citation field extraction\\
\hline
~\cite{jagannatha2016structured} & $\backslash$ & Trained with skip-gram & $\backslash$ & Bi-LSTM &
Skip-chain CRF & Clinical entities detection \\
\hline
~\cite{wu2018evaluating} & Word shapes, gaz & Glove & CNN & Bi-LSTM & CRF & NER \\
\hline
~\cite{Collobert2011Natural} & Gaz, cap & Senna & $\backslash$ & CNN & CRF & POS, NER, chunking, SRL \\
\hline
~\cite{wang2017named} & $\backslash$ & Glove & CNN & Gated-CNN & CRF & NER \\
\hline
~\cite{strubell2017fast} & $\backslash$ & Word2vec & $\backslash$ & ID-CNN & CRF & NER \\
\hline
~\cite{Lample2016Neural} & $\backslash$ & Word2vec & Bi-LSTM & Bi-LSTM & CRF & NER \\
\hline
~\cite{huang2015bidirectional} & Spelling, gaz & Senna & $\backslash$  & Bi-LSTM & CRF & POS, NER, chunking \\
\hline
~\cite{Santos2014Learning} & $\backslash$ & Word2vec & CNN & CNN & CRF & POS\\
\hline
~\cite{zhai2017neural} & $\backslash$ & Senna & CNN & Bi-LSTM & Pointer network & Chunking, slot filling \\
\hline
~\cite{zheng2017joint} & $\backslash$ & Word2vec & $\backslash$ & Bi-LSTM & LSTM & Entity relation extraction\\
\hline
~\cite{ghaddar2018robust} & LS vector, cap & SSKIP & Bi-LSTM & LSTM & CRF & NER \\
\hline
~\cite{shen2017deep} & $\backslash$ & Word2vec & CNN & CNN & LSTM & NER \\
\hline
~\cite{li2018segbot} & $\backslash$ & Glove & $\backslash$ & Bi-GRU & Pointer network & Text segmentation \\
\hline
~\cite{gui2017part} & $\backslash$ & - & CNN & Bi-LSTM & Softmax & POS\\
\hline
~\cite{dozat2017stanford} & $\backslash$ & Word2vec, FastText & LSTM+attention & Bi-LSTM & Softmax & POS \\
\hline
~\cite{hu2019neural} & $\backslash$ & Glove & CNN & Bi-LSTM & NCRF transducers & POS, NER, chunking\\
\hline
~\cite{kann2018character} & $\backslash$ & - & Bi-LSTM+AE & Bi-LSTM & softmax & POS \\
\hline
~\cite{sato2017segment} & Lexicons & Glove & CNN & Bi-LSTM & Segment-level CRF & NER \\
\hline
~\cite{chen2019grn} & $\backslash$ & Glove & CNN & GRN+CNN & CRF & NER \\
\hline
~\cite{wei2020position} & $\backslash$ & Glove & CNN & Bi-LSTM+SA & CRF & POS, NER, chunking
\\
\hline
		\end{tabular}}
\label{table_summary}

\end{table*}

In this section, we survey deep learning based approaches for sequence labeling. We present the review with a scientific taxonomy that categorize existing works along three axes: embedding module, context encoder module, and inference module, of which three stages neural sequence labeling models often consists. The embedding module is the first stage that maps words into their distributed representations. The context encoder module extracts contextual features and the inference module predict labels and generate optimal label sequence as output of the model.
In Table \ref{table_summary}, we make a brief overview of the deep learning based sequence labeling models with the aforementioned taxonomy. We list the different architectures that these work adopt in the three stages and the final column give the focused tasks.

\subsection{Embedding Module}
The embedding module maps words into their distributed representations as the initial input of model. An embedding lookup table is usually required to convert the one-hot encoding of each word to a low dimensional real-valued dense vector, where each dimension represents a latent feature.
In addition to pretrained word embeddings, character-level representations, hand-crafted features and sentence-level representations can also be part of the embedding module, supplementing features for the initial input from different perspectives.

\subsubsection{Pretrained Word Embeddings}
Pretrained word embeddings that learned on a large corpus of unlabeled data has become a key component in many neural NLP models. Adopting it to initialize the embedding lookup table can achieve significant improvements over randomly initialized ones, since the syntactic and semantic information within language are captured during the pretraining process. There are many published pretrained word embeddings that have been widely used, such as Word2Vec, Senna, GloVe and \etc

Word2vec~\cite{mikolov2013efficient} is a popular method to compute vector representations of words, which provides two model architectures including the continuous bag-of-words and skip-gram. Santos et al.~\cite{Santos2014Learning} use word2vec's skip-gram method to train word embeddings and added to their sequence labeling model. Similarly, some work~\cite{Rei2017Semi,rei2016attending,dozat2017stanford} initialize the word embeddings in their model with publicly available pretrained vectors that created using word2vec. Lample et al.~\cite{Lample2016Neural} apply skip-n-gram~\cite{ling2015not} to pretrain their word embeddings, which is a variation of word2vec that accounts for word order. Gregoric et al.~\cite{gregoric2018named} follow their work and also used the same embeddings.

Collobert et al.~\cite{Collobert2011Natural} propose the ``SENNA" architecture in 2011, which pioneers the idea of solving natural language processing tasks from the perspective of neural language model, and it also includes a construction method of pretrained word embeddings. Many subsequent work~\cite{Peters2017Semi,Zhang2017Does,chiu2016named,huang2015bidirectional,yang2016multi,ma2016new} adopt SENNA word embeddings as the initial input of their sequence labeling models.
Besides, Stanford's publicly available GloVe embeddings~\cite{pennington2014glove} that trained on $6$ billion token corpus from Wikipedia and web text are also widely used and adopted by many work
~\cite{Ma2016End,Yasunaga2018Robust,Liu2017Empower,akbik2018contextual,ye2018hybrid,xin2018learning,li2018segbot,hu2019neural,chen2019grn} to initilize their word embeddings.

The above pretrained word embedding methods only generate a single context-independent vector for each word, ignoring the modeling of polysemy problem. Recently, many approaches for learning contextual word representations~\cite{Peters2017Semi,Peters2018Deep,akbik2018contextual} have been proposed, where bidirectional language models (LM) are trained on a large unlabeled corpus and the corresponding internal states are utilized to produce a word representation. And the representation of each word is dependent on its context. For instance, the generated embedding of the word ``present" in ``How many people were present at the meeting?" is different from that in ``I'm not at all satisfied with the present situation".

Peters et al.~\cite{Peters2017Semi} propose pretrained contextual embeddings from bidirectional language models and added them to sequence labeling model, achieving pretty excellent performance on the task of NER and chunking. The method first pretrains the forward and backward neural language model separately with Bi-LSTM architecture on a large, unlabeled corpus. Then it removes the top softmax layer and concatenates the forward and backward LM embeddings to form bidirectional LM embeddings for every token in a given input sequence.
Peters et al. extend their method~\cite{Peters2017Semi} in ~\cite{Peters2018Deep} by introducing ELMo (Embeddings from Language Models) representations. Unlike previous approaches that just utilize the top LSTM layer, the ELMo representations are a linear combination of internal states of all bidirectional LM layers, where the weight of each layer is task-specific. By adding these representations to existing models, the method significantly improves the performance across a broad range of diverse NLP tasks~\cite{chen2019grn,hu2019neural,clark2018semi}.

 Akbik et al.~\cite{akbik2018contextual} propose a similar method to generate pretrained contextual word embeddings by adopting a bidirectional character-aware language model, which learns to predict the next and previous character instead of word.

Devlin et al.~\cite{devlin2018bert} propose a pretraining language representation model called BERT, which stands for Bidirectional Encoder Representations from Transformers. It obtains new state-of-the-art results on eleven tasks and causes a sensation in the NLP communities. The core idea of BERT is to pretrain deep bidirectional representations by jointly conditioning on both left and right context in all layers.
Although the sequence labeling tasks can be addressed by fine-tuning the existing pre-trained BERT model,
the output hidden states of BERT can also be taken as additional word embeddings to promote the performance of sequence labeling models~\cite{liu2019gcdt,luo2020hierarchica}.

By modeling the context information, the word representations produced by ELMo and BERT can encode rich semantic information.
In addition to such context modeling, a recent work proposed by He et al.~\cite{he2020knowledge-graph} provide a new kind of word embedding that is
both context-aware and knowledge-aware, which encode the prior knowledge of entities from an external knowledge base. The proposed knowledge-graph augmented word representations significantly promotes the performance of NER in various domains.

\subsubsection{Character-level Representations}
Although syntactic and semantic information are captured inside the pretrained word embeddings, the word morphological and shape information is normally ignored, which is extremely useful for many sequence labeling tasks like part-of-speech tagging. Recently,
many researches learn the character-level representations of words through neural networks and incorporate them into the embedding module of models to exploit useful intra-word information, which can also tackle the out-of-vocabulary word problem effectively and has been verified to be helpful in numerous sequence labeling tasks. The two most common architectures to capture character-to-word representations are Convolutional Neural Networks(CNNs) and Recurrent Neural Networks(RNNs).

\paratitle{Convolutional Neural Networks}.

\begin{figure}[!t]
	\centering
	\includegraphics[width=0.4\textwidth]{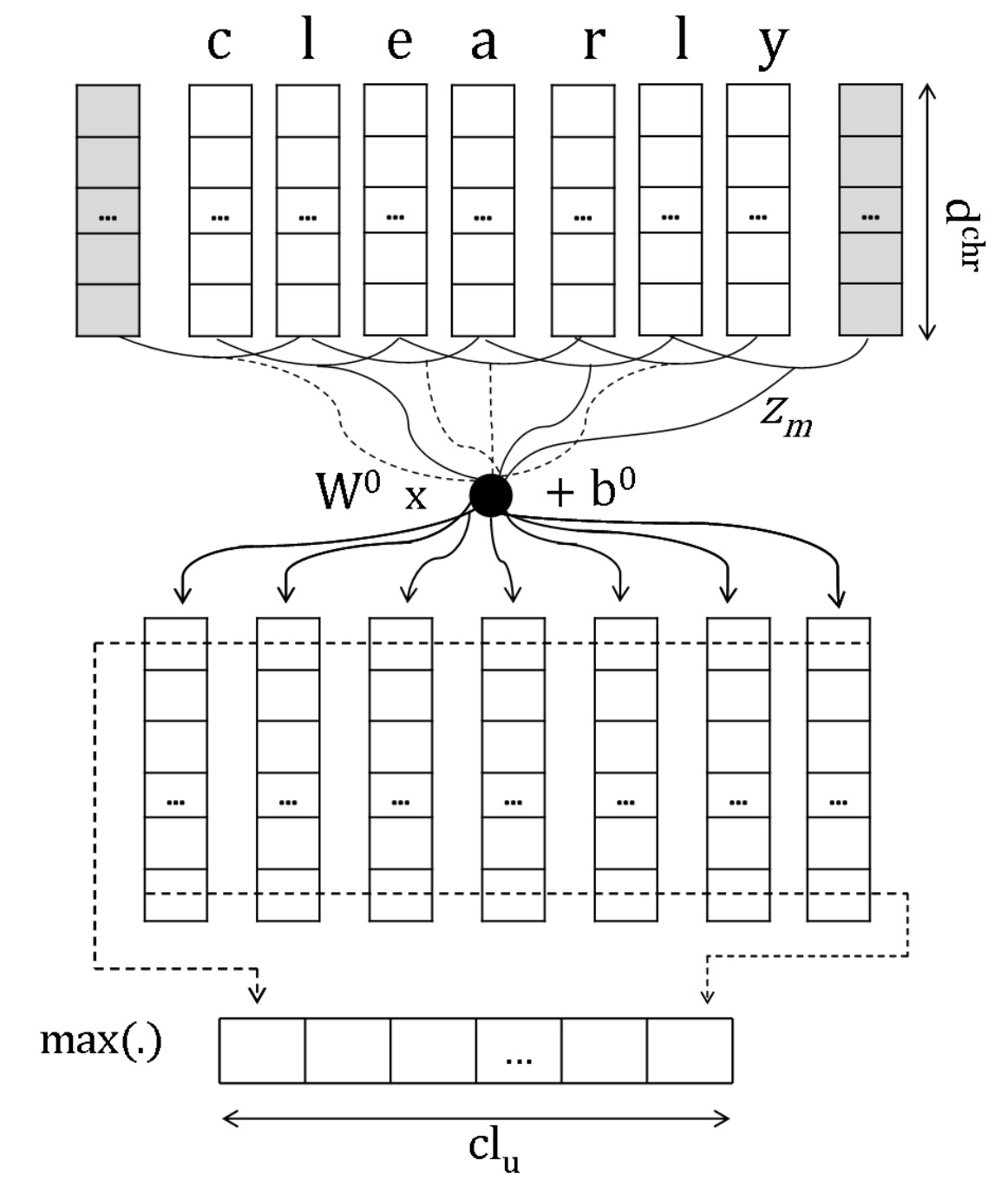}
	\caption{Convolutional approach to character-level feature extraction~\cite{Santos2014Learning}.}
	\label{fig:42}
\end{figure}

Santos and Zadrozny~\cite{Santos2014Learning} initially propose the approach that using CNNs to learn character-level representations of words for sequence labeling, which is followed by many subsequent work~\cite{Ma2016End,chiu2016named,shen2017deep,chen2019grn,wu2018evaluating,clark2018semi,Zhao2019ANM}. The approach applies a convolutional operation to the sequence of character embeddings and produces local features of each character. Then a fixed-sized character-level embedding of the word is extracted by using the max over all character windows. The process is depicted in Fig \ref{fig:42}.

\begin{figure}[!t]
	\centering
	\includegraphics[width=0.4\textwidth]{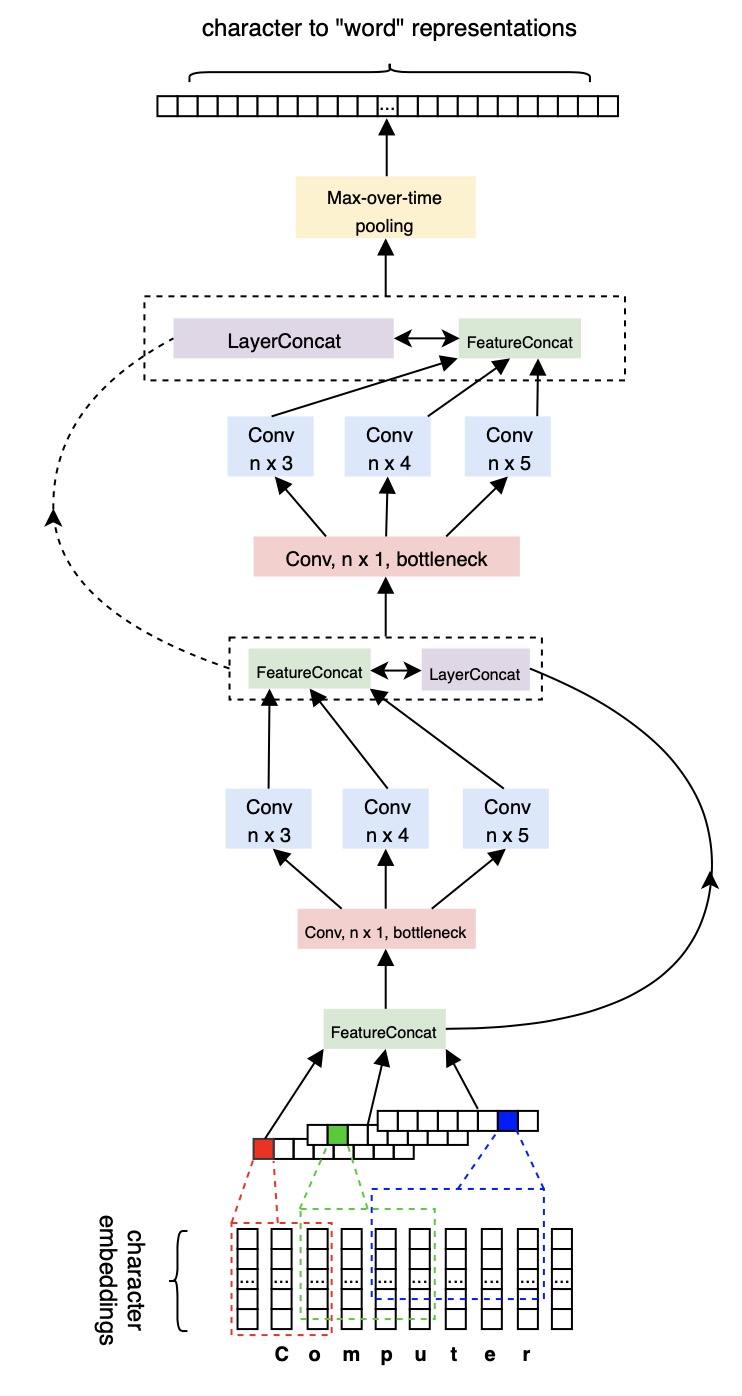}
	\caption{The main architecture of IntNet~\cite{xin2018learning}.}
	\label{fig:28}
\end{figure}

Xin et al.~\cite{xin2018learning} propose IntNet, a funnel-shaped wide convolutional neural network for learning character-level representations for sequence labeling. Unlike previous CNN-based character embedding approaches, this method delicately designs the convolutional block that comprises of several consecutive operations, and utilizes multiple convolutional layers in which feature maps are concatenated in every other ones. It helps the network to capture different levels of features and explore the full potential of CNNs to learn better internal structure of words. The proposed model achieves significant improvements over other character embedding models and obtains state-of-the-art performance on various sequence labeling datasets. Its main architecture can be shown in Fig \ref{fig:28}.

\paratitle{Recurrent Neural Networks}.

\begin{figure}[!t]
	\centering
	\includegraphics[width=0.4\textwidth]{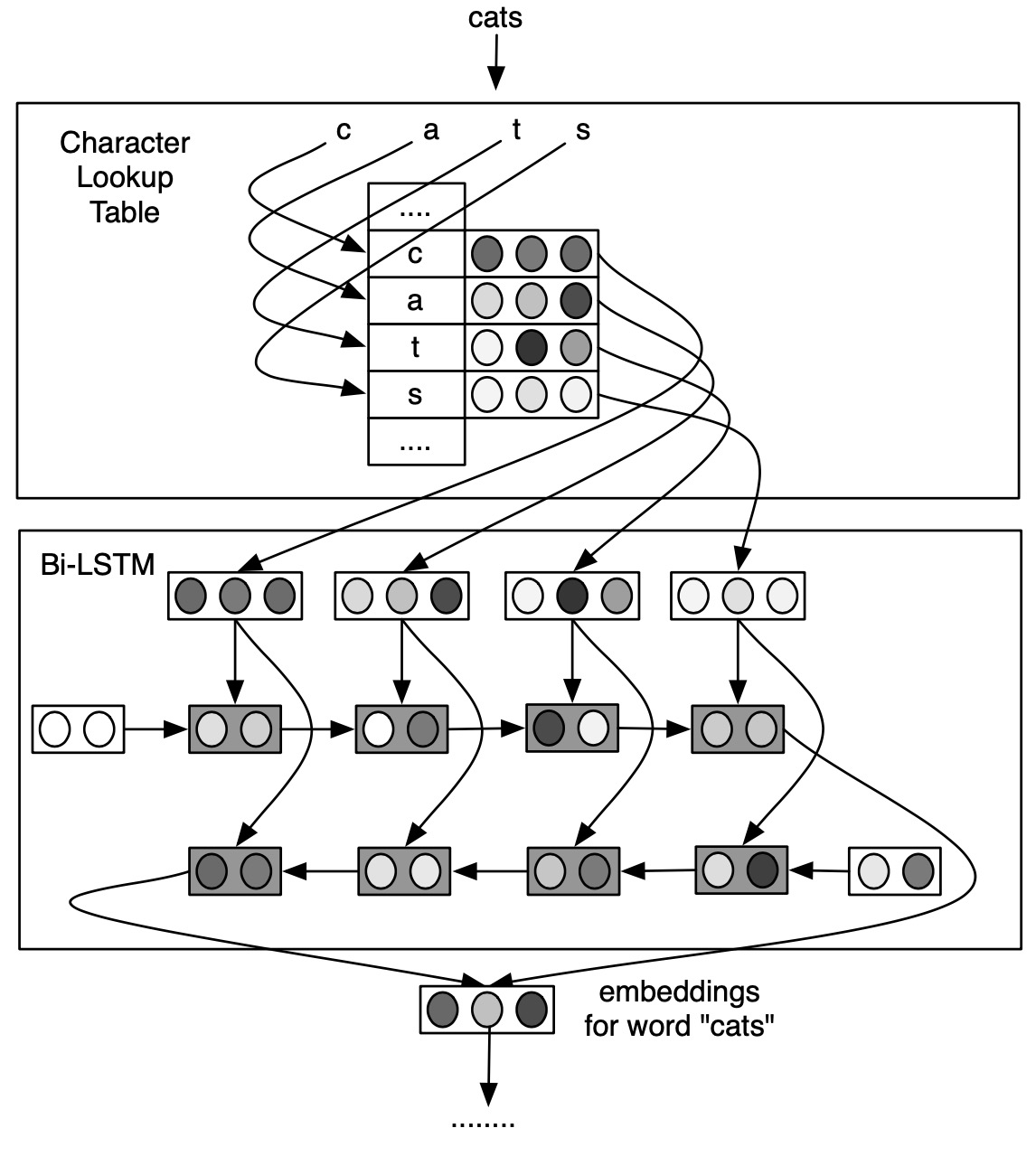}
	\caption{Illustration of the lexical Composition Model~\cite{ling2015finding}.}
	\label{fig:25}
\end{figure}

Ling et al.~\cite{ling2015finding} propose a compositional character to word (C2W) model that uses bidirectional LSTMs (Bi-LSTM) to build word embeddings by taking the characters as atomic units. A forward and a backward LSTM processes the character embeddings sequence of a word in direct and reverse order. And the representaion for a word derived from its characters is obtained by combining the final states of the bidirectional LSTM. Illustration of the proposed method is shown in Fig \ref{fig:25}. By exploiting the 
features in language effectively,
the C2W model yields excellent results in language modeling and part-of-speech tagging. And many work~\cite{Lample2016Neural,zhang2018learning,gregoric2018named,Yasunaga2018Robust,plank2016multilingual} follow them to apply Bi-LSTM for obtaining character-level representations for sequence labeling.
Similarly, Yang et al.~\cite{Yang2016Transfer} employ GRUs for the character embedding model instead of LSTM units.

Dozat et al.~\cite{dozat2017stanford} propose a RNN based character-level model in which the character embeddings sequence of each word is fed into a unidirectional LSTM followed by an attention mechanism. The method first extracts the hidden state and cell state of each character from the LSTM and then computes linear attention over the hidden states. The output of attention is concatenated with cell state of the final character to form the character-level word embedding for their POS tagging model.

Kann et al.~\cite{kann2018character} propose a character-based recurrent sequence-to-sequence architecture, which connects the Bi-LSTM character encoding model to a LSTM based decoder that associated with an auxiliary objective (random string autoencoding, word autoencoding or lemmatization). The multi-task architecture introduces additional character-level supervision into the model, which helps them build a more robust neural POS taggers for low-resource languages.

Bohnet et al.~\cite{bohnet2018morphosyntactic} propose a novel sentence-level character model for learning context sensitive character-based representations of words. Unlike all aforementioned token-level character model, this method feeds all characters of a sentence into a Bi-LSTM layer and concatenates the forward and backward output vector of the first and last character in the word to form its final character-level representation. This strategy allows context information to be incorporated in the initial word embeddings before flowing into the context encoder module. Similarly, Liu et al.~\cite{Liu2017Empower} also adopt the character-level Bi-LSTM that processes all characters of a sentence instead of a word. However, their proposed model focuses on extracting knowledge from raw texts by leveraging the neural language model to effectively extract character-level information. In particular, the forward and backward character-level LSTM would predict the next and previous word at word boundaries. In order to mediate the primary sequence labeling task and the auxiliary language model task, highway networks are further employed, which transform the output of the shared character-level layer into two different representations. One is used for language model and the other can be viewed as character-level representation that combined with the word embedding for sequence labeling model.

\subsubsection{Hand-crafted features}

As aforementioned, enabled by the powerful capacity to extract features automatically, deep neural network based models have the advantage of not requiring complex feature engineering. However, before fully end-to-end deep learning models~\cite{Ma2016End,Lample2016Neural} are proposed for sequence labeling tasks, feature engineering is typically utilized in neural models~\cite{Collobert2011Natural,huang2015bidirectional,chiu2016named}, where hand-crafted features such as word spelling features that can greatly benefits POS tagging and gazetteer features that are widly used in NER are represented as discrete vectors and then integrated to the embedding module. For example, Collobert et al.~\cite{Collobert2011Natural} utilize word suffix, gazetteer and capitalization features as well as cascading features that include tags from related tasks. Huang et al.~\cite{huang2015bidirectional} adopt designed spelling features (include word prefix and suffix features, capitalization feature \etc), context features (unigram, bi-gram and tri-gram features) and gazetteer features. Chiu and Nichols~\cite{chiu2016named} use character-type, capitalization, and lexicon features.

In recent two years, there have been some work~\cite{wu2018evaluating,ghaddar2018robust,rijhwani2020soft,lin2019gazetteer,liu2019towards} that focus on incorporating manual features into neural models in a more effective manner and obtain significant further improvements for sequence labeling. Wu et al.~\cite{wu2018evaluating} propose a hybrid neural model which combines a feature auto-encoder loss component to utilize hand-crafted features, and significantly outperforms existing competitive models on the task of NER. Exploited manual features include part-of-speech tags, word shapes and gazetteers. In particular, the auto-encoder auxiliary component takes hand-crafted features as input and learns to re-construct them into output, which helps the model to preserve important information stored in these features and thus enhances the primary sequence labeling task.
Their proposed method has demonstrated the utility of hand-crafted features for named entity recognition on English data. However, designing such features for low-resource languages is challenging, because gazetteers in these languages are absent.
To address this proplem,
Rijhwani et al.~\cite{rijhwani2020soft} propose a method of `"soft gazetteers"
that incorporates information from English knowledge bases through cross-lingual entity linking and create continuous-valued gazetteer features for low-resource languages.

Ghaddar et al.~\cite{ghaddar2018robust} propose a novel lexical representation (called Lexical Similarity \ie (LS) vector) for NER, indicating that robust lexical features are quiet useful and can greatly benefit deep neural network architectures. The method first embeds words and named entity types into a joint low-dimensional vector space, which is trained from a Wikipedia corpus annotated with 120 fine-grained entity types. Then a 120-dimensional feature vector (\ie LS vector) for each word is computed offline, where each dimension encodes the similarity of the word embedding with the embedding of an entity type. The LS vectors are finally incorporated into the embedding module of their neural NER model.

\subsubsection{Sentence-level Representations}

\begin{figure}[t!]
	\centering
	\includegraphics[width=0.5\textwidth]{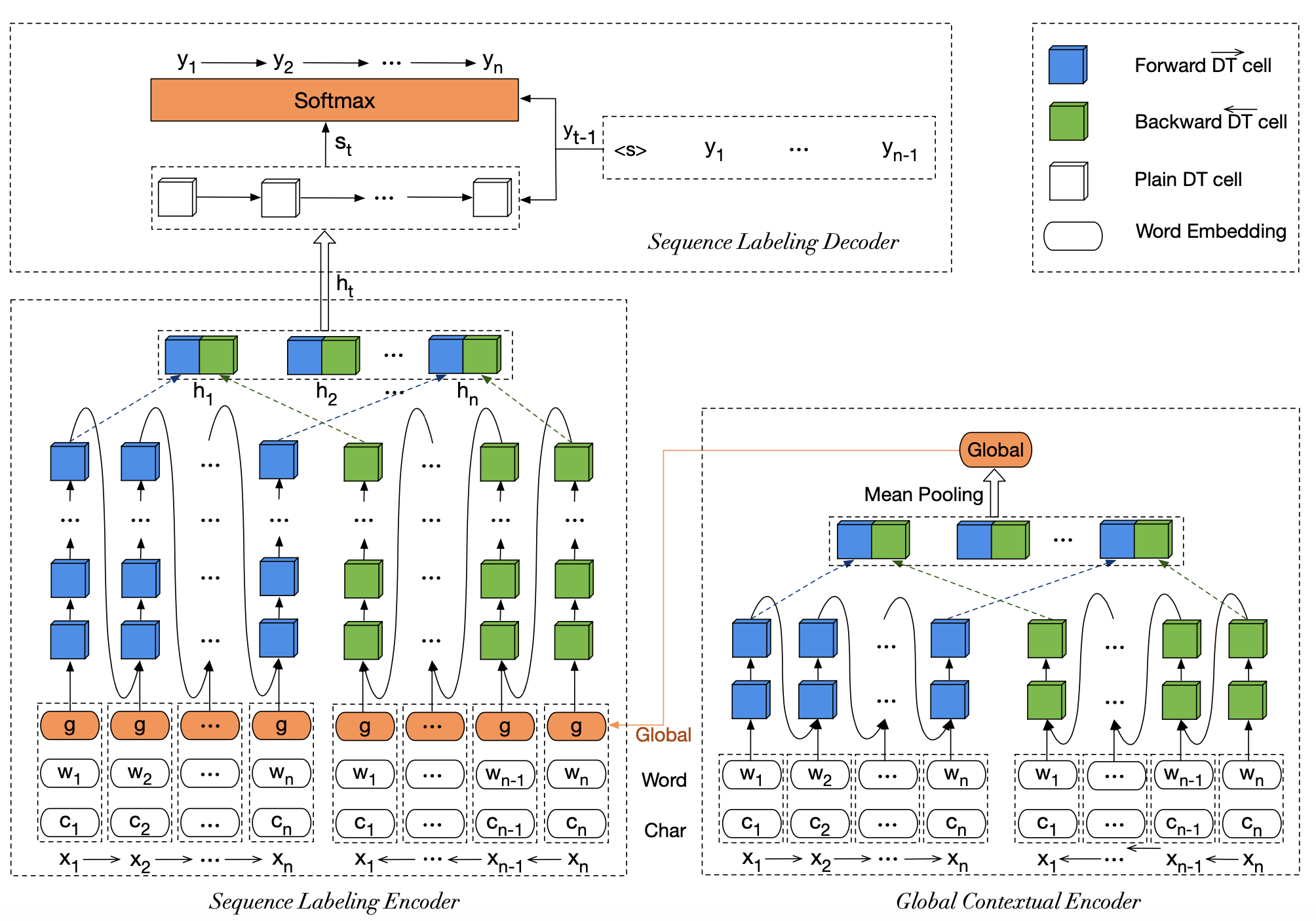}
	\caption{The global contextual encoder (on the right) outputs sentence-level representation that serves as an enhancement of token representation.~\cite{liu2019gcdt}.}
	\label{fig:gcdt}
\end{figure}

Existing research~\cite{zhang2018sentence} has proved that the global contextual information from the entire sentence is useful for modeling sequence, which is insufficiently captured at each token position in context encoder like Bi-LSTM.
To solve this problem, some recent work~\cite{liu2019gcdt,luo2020hierarchical} have introduced sentence-level representations into the embedded module, that is, in addition to pretrained word embeddings and character-level representations, they also assign every word with a global representation learned from the entire sentence, which can be shown in Fig \ref{fig:gcdt}.
Though these work propose different ways to get sentence representation,
they all prove the superiority of adding it in the embedding module for the final performance of sequence labeling tasks.

\subsection{Context Encoder Module}
Context dependency plays a significant role in sequence labeling tasks. The context encoder module extracts contextual features of each token and capture the context dependencies of given input sequence. Learned contextual representations will be passed into inference module for label prediction.
There are three commonly used model architectures for context encoder module, \ie RNN, CNN and Transformers.

\subsubsection{Recurrent Neural Network}
Bi-LSTM is almost the most widely used context encoder architecture today. Concretely, it incorporates past/future contexts from both directions (forward/backward) to generate the hidden states of each token, and then jointly concatenate them to represent the global information of the entire sequence. While Hammerton~\cite{hammerton2003named} has studied utilizing LSTMs for NER tasks in the past, the lack of computing power limits the effectiveness of their model. With recent advances in deep learning, much research effort has been dedicated to using Bi-LSTM architecture and achieve excellent performance. Huang et al.~\cite{huang2015bidirectional} initially adopt Bi-LSTM to generate contextual representations of every word in their sequence labeling model, and produce state-of-the-art accuracy on POS tagging, chunking and NER data sets. Similarly, ~\cite{zhang2018learning,ling2015finding,Ma2016End,Lample2016Neural,chiu2016named,bohnet2018morphosyntactic,Yasunaga2018Robust,Liu2017Empower,plank2016multilingual} also choose the same Bi-LSTM architecture for context encoding. Gated Recurrent Unit(GRU) is a variant of LSTM which also addresses long-dependency issues in RNN networks, and several work utilize Bi-GRU as their context encoder architecture~\cite{Yang2016Transfer,li2018segbot,zhu2019can}.

Rei~\cite{Rei2017Semi} propose a multitask learning method that equips the Bi-LSTM context encoder module with a auxiliary training objective, which learns to predict surrounding words for every word in the sentence. It shows that the language modeling objective provides consistent performance improvements on several sequence labeling benchmark, because it motivates the model to learn more general semantic and syntactic composition patterns of the language.

Zhang et al.~\cite{Zhang2017Does} propose a new method called Multi-Order BiLSTM which combines low order and high order LSTMs together in order to learn more tag dependencies. The high order LSTMs predict multiple tags for the current token which contains not only the current tag but also the previous several tags. The model keeps the scalability to high order models with a pruning technique, and achieves the state-of-the-art result in chunking and highly competitive results in two NER datasets.

Ma et al.~\cite{ma2017jointly} propose a LSTM-based model for jointly training sentence-level classification and sequence labeling tasks, in which a modified LSTM structure is adopted as their context encoder module. In particular, the method employs a convolutional neural network before LSTM to extract features from both the context and previous tags of each word. Therefore, the input for LSTM is changed to include meaningful contextual and label information.

Most of the existing LSTM based methods use one or more stacked LSTM layers to extract context features of words. However, Gregoric et al.~\cite{gregoric2018named} present a different architecture which employs multiple parallel independent Bi-LSTM units across the same input and promotes diversity among them by employing an inter-model regularization term. It shows that the method reduces the total number of parameters in the model and achieves significant improvements on the CoNLL 2003 NER dataset compared to other previous methods.

Kazi et al.~\cite{kazi2017implicitly} propose a novel implicitly-defined neural network architecture for sequence labeling. In contrast to traditional recurrent neural networks, this work provides a different mechanism that each state is able to consider information in both directions. The method extends RNN by changing the definition of implicit hidden layer function:
$$
h_t = f(\xi_t,h_{t-1},h_{t+1})
$$
where $\xi_t$ denotes the input of hidden layer, $h_{t-1}$ and $h_{t+1}$ is the hidden state of last and next time step, respectively. It forgoes the causality assumption used to formulate RNN and leads to an implicit set of equations for the entire sequence of hidden states. They compute them via an approximate Newton solve and apply the Krylov Subspace method~\cite{knoll2004jacobian}. The implicitly-defined neural network architecture helps to achieve improvements on problems with complex, long-distance dependencies.

Although Bi-LSTM has been widely adopted as context encoder architecture,
there are still several natural limitations, such as the shallow connections
between consecutive hidden states of RNNs.
At each time step, BiLSTMs consume an incoming word and construct a new summary of the past subsequence. This process should be highly non-linear so that the hidden states can quickly adapt to variable inputs while still retaining useful summaries of the past~\cite{pascanu2013construct}.
Deep transition RNNs extend conventional RNNs by increasing the transition depth of consecutive hidden states~\cite{pascanu2013construct}.
Recently, Liu et al.~\cite{liu2019gcdt} introduce the deep transition architecture
for sequence labeling and achieve a significant performance improvement on the tasks of
text chunking and NER.
Besides, the way of sequentially processing inputs of RNN might limit the ability to capture the non-continuous relations over tokens within a sentence. To tackle the problem, a recent work proposed by Wei et al.~\cite{wei2020position} employs self-attention to provide complementary context information on the basis of Bi-LSTM. They propose a position-aware self-attention as well as a well-designed self-attentional context fusion network, aiming to explore the relative positional information of an input sequence for capturing the latent relations among tokens. It shows that the method achieves significant improvements on the tasks of POS, NER and chunking.

\subsubsection{Convolutional Neural Networks}

Convolutional Neural Networks (CNNs) are another popular architecture for encoding context information in sequence labeling models. Compared to RNN, CNN based methods are considerably faster since it can fully leverage the GPU parallelism through the feed-forward structure. An initial work in this area is proposed by Collobert et al.~\cite{Collobert2011Natural}. The method employs a simple feed-forward neural network with a fixed-size sliding window over the input sequence embedding, which can be viewed as a simplified CNN without pooling layer. And this window approach is based on the assumption that the label of a word depends mainly on its neighbors. Santos et al.~\cite{Santos2014Learning} follow their work and use similar structure for context feature extraction.

Shen et al.~\cite{shen2017deep} propose a deep active learning based model for NER tasks. Their tagging model extracts context representations for each word using a CNN due to its strong efficiency, which is crucial for their iterative retraining scheme. The structure has two convolutional layers with kernels of width three, and it concatenates the representation at the last convolutional layer with the input embedding to form the output.

Wang et al.~\cite{wang2017named} employ stacked Gated Convolutional Neural Networks(GCNN) for named entity recognition, which extend the convolutional layer with gating mechanism. In particular, a gated convolutional layer can be written as
$$
F_{gating}(\matrix{X}) = (\matrix{X} * \matrix{W}+\hat{b})\odot \sigma(\matrix{X}* \matrix{V}+\hat{c})
$$
where $*$ denotes row convolution, $\matrix{X}$ is the input of this layer, $\matrix{W},\hat{b},\matrix{V},\hat{c}$ are the parameters to be learned, $\sigma$ is the sigmoid function and represents element-wise product.

Though relatively high efficiency, a major disadvantage of CNNs is that it has difficulties in capturing long-range dependencies in sequences due to the limited receptive fields, which makes fewer methods to perform sequence labeling tasks with CNNs than RNNs. In recent year, some CNN-based models modify traditional CNNs to better capture global context information and achieve excellent results for sequence labeling.

\begin{figure}[!t]
	\centering
	\includegraphics[width=0.5\textwidth]{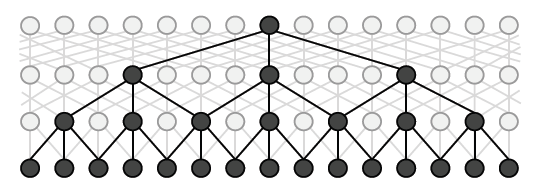}
	\caption{A dilated CNN block with maximum dilation width 4 and filter width 3. Neurons contributing to a single highlighted neuron in the last layer are also highlighted~\cite{strubell2017fast}.}
	\label{fig:38}
\end{figure}

Strubell et al.~\cite{strubell2017fast} propose a Iterated Dilated Convolutional Neural Networks (ID-CNNs) method for the task of NER, which enables significant speed improvements while maintaining accuracy comparable to the state-of-the-arts. Dilated convolutions~\cite{yu2015multi} operate on a sliding window of context like typical CNN layers, but the context need not be consecutive. The convolution is defined over a wider effective input width by skipping over several inputs at a time, and the effective input width can grow exponentially with the depth. Thus it can incorporate broader context into the representation of a token than typical CNN. Fig \ref{fig:38} shows the structure.

The proposed iterated dilated CNN architecture repeatedly applies the same block of dilated convolutions to token-wise representations. Repeatedly employing the same parameters prevents overfitting problem and provides the model desirable generalization capabilities.

Chen et al.~\cite{chen2019grn} propose gated relation network(GRN) for NER, in which a gated relation layer that models the relationship between any two words is built on top of CNNs for capturing long-range context information. Specifically, it firstly computes the relation score vector between any two words,
$$
r_{ij} = \matrix{W}_{rx}[x_i;x_j] + b_{rx}
$$
where $x_i$ and $x_j$ denote the local context features from the CNN layer for the i-th and j-th word in the sequence, $\matrix{W}_{rx}$ is the weight matrix and $b_{rx}$ is the bias vector. Note that the relation score vector $r_{ij}$ is of the same dimension as $x_i$ and $x_j$. Then the corresponding global contextual representation $r_i$ for the ith word is obtained by performing a weighted-summing up operation, in which a gating mechanism is adopted for adaptively selecting other dependent words.
$$
r_i = \frac{1}{T}\sum_{j=1}^T\sigma(r_{ij})\odot x_j
$$
where $\sigma$ is a gate using sigmoid function, and $\odot$ denotes element-wise multiplication. The proposed GRN model achieves significantly better performance than ID-CNN~\cite{strubell2017fast}, owing to its stronger capacity to capture global context dependencies.

\subsubsection{Transformers}
The Transformer model is proposed by Vaswani et al.~\cite{Vaswani2017Attention} in 2017 and achieves excellent performance for Neural Machine Translation (NMT) tasks. The overall architecture is based solely on attention mechanisms to draw global dependencies between input, dispensing with recurrence and convolutions entirely. The initial proposed Transformer employs a sequence to sequence structure that comprises the encoder and decoder. But the subsequent research work often adopt the encoder part to serve as the feature extractor, thus our introduction here is limited to it.


The encoder is composed of a stack of several identical layers, which includes a multi-head self-attention mechanism and a position-wise fully connected feed-forward network. It employs a residual connection~\cite{he2016deep} around each of the two sub-layers to ease the training of deep neural network. And layer normalization~\cite{lei2016layer} is applied after the residual connection to stabilize the activations of model.

Due to its superior performance, the Transformer is widely used in various NLP tasks and has achieved excellent results. However, in sequence labeling tasks, the Transformer encoder has been reported to perform poorly~\cite{guo2019star}.
Recently, Yan et al.~\cite{yan2019tener} analyze the properties of Transformer for
exploring the reason why Transformer does not work well in sequence labeling tasks especially NER.
Both the direction and relative distance information are important in the NER, but
these information will lose when the sinusoidal position embedding is used in the vanilla Transformer.
To address the problem,
they
propose TENER, an architecture adopting adapted Transformer Encoder by incorporating the direction and relative distance aware attention and the un-scaled attention,
which can greatly boost the performance of Transformer encoder for NER.
Star-Transformer is a lightweight alternative of Transformer proposed by Shao et al.~\cite{guo2019star}. It replaces the fully-connected structure with a star-shaped topology, in which every two non-adjacent nodes are connected through a shared relay node. The model complexity is reduced significantly, and it also achieved great improvements against the standard Transformer on various tasks including sequence labeling tasks.

\subsection{Inference Module}
The inference module takes the representations from context encoder module as input, and generate the optimal label sequence.

\subsubsection{Softmax}
The softmax function that also called normalized exponential function, is a generalization of logic functions and has been widely used in a variety of probability-based multi-classification methods. It maps a $K $-dimensional vector $z$ into another $K$-dimensional real vector $\sigma(z)$ such that each element has a range between $0$ and $1$ and the sum of all elements equals $1$. The form of the function is usually given by the following formula
$$
\sigma(z)_j = \frac{e^{z_j}}{\sum_{k=1}^Ke^{z_k}}
$$
where $j=1,\dots,K$.

Many models for sequence labeling treat the problem as a set of independent classification tasks, and utilize a softmax layer as a linear classifier to assign optimal label for each word in a sequence~\cite{bohnet2018morphosyntactic,gregoric2018named,ling2015finding,kazi2017implicitly,gui2017part,ma2017jointly,kemos2018neural,kim2017cross}. Specifically, given the output representation $h_t$ of the context encoder at time step $t$, the probability distribution of the t-th word's label can be obtained by a fully connected layer and a final softmax function
$$
o_t = softmax(\matrix{W}h_t+b)
$$
where the weight matrix $\matrix{W}\in{R^{d\times|T|}}$ maps $h_t$ to the space of labels, $d$ is the dimension of $h_t$ and $|T|$ is the number of all possible labels.

\subsubsection{Conditional Random Fields}
The above methods of independently inferring word labels in a given sequence ignore the dependencies between labels. Typically, the correct label to each word often depends on the choices of nearby elements. Therefore, it is necessary to consider the correlation between labels of adjacent neighborhoods to jointly decode the optimal label chain of the entire sequence. CRF model~\cite{knoll2004jacobian} has been proven to be powerful in learning the strong dependencies across output labels, thus most of the neural network-based models for sequence labeling employ CRF as the inference module~\cite{Ma2016End,Yasunaga2018Robust,Liu2017Empower,plank2016multilingual,Rei2017Semi,Peters2017Semi,akbik2018contextual,zhang2017embracing,cao2018adversarial,feng2018improving}.

Specifically,
%
let $\matrix{Z} = [\hat{z}_{1},\hat{z}_{2}, \ldots,\hat{z}_{n}]^{\top}$
be the output of context encoder of the given sequence $\hat{x}$,
the probability $\Pr(\hat{y}|\hat{x})$ of generating the whole label sequence $y_i\in\hat{y}$ with regard to $\matrix{Z}$
is
$$
  \Pr(\hat{y}|\hat{x}) = \frac{\prod_{j=1}^{n}\phi(y_{j-1},y_j,\hat{z}_j)}
{\sum_{y^{'}\in{\matrix{Y}(\matrix{Z})}}\prod_{j=1}^n\phi(y^{'}_{j-1},y^{'}_j,\hat{z}_j)},
$$
where $\matrix{Y}(\matrix{Z})$ is the set of possible label sequences for $\matrix{Z}$;
\begin{math}
  \phi(y_{j-1},y_j,\hat{z}_j)\!=\!\exp(\matrix{W}_{y_{j-1},y_{j}}\hat{z}_j + b_{y_{j-1},y_j}),
\end{math}
$\matrix{W}_{y_{j-1},y_{j}}$ and  $b_{y_{j-1},y_j}$
indicate the weighted matrix and bias parameters corresponding to the label pair $(y_{j-1},y_j)$, respectively.

\paratitle{Semi-CRF}.

Semi-Markov conditional random fields (semi-CRFs)~\cite{sarawagi2005semi} is an extension of conventional CRFs, in which labels are assigned to the segments of input sequence rather than to individual words. It extracts features of segments and models the transition between them, suitable for segment-level sequence labeling tasks such as named entity recognition and phrase chunking. Compared to CRFs, the advantage of semi-CRFs is that it can make full use of segment-level information to capture the internal properties of segments, and higher-order label dependencies can be taken into account. However, since it jointly learns to determine the length of each segment and the corresponding label, the time complexity becomes higher. Besides, more features is required for modeling segments with different lengths and automatically extracting meaningful segment-level features is an important issue for Semi-CRFs. With advances in deep learning, some models combining neural networks and Semi-CRFs for sequence labeling have been studied.

Kong et al.~\cite{kong2015segmental} propose Segmental Recurrent Neural Networks (SRNNs) for segment-level sequence labeling problems, which adopts a semi-CRF as the inference module and learns representations of segments through Bi-LSTM. Based on the recurrent nature of RNN, this method further designs a dynamic programming algorithm to reduce the time complexity. A parallel work Gated Recursive Semi-CRFs (grSemi-CRFs) proposed by Zhuo et al.~\cite{zhuo2016segment} employs a Gated Recursive Convolutional Neural Network (grConv)~\cite{cho2014properties} to extract segment features for semi-CRF. The grConv is a variant of recursive neural network that learns segment-level representations by constructing a pyramid-like structure and recursively combining adjacent segment vectors. The follow-up work proposed by Kemos et al.~\cite{kemos2018neural} utilize the same grConv architecture for extracting segment features in their neural semi-CRF model. It takes characters as the basic input unit but does not require any correct token boundaries, which is different from existing character-level models. The model is based on semi-CRF to jointly segment (tokenize) and label characters, being robust for languages with difficult or noisy tokenization. Sato et al.~\cite{sato2017segment} design Segment-level Neural CRF for segment-level sequence labeling tasks. The method applies a CNN to obtain segment-level representations and constructs segment lattice to reduce search space.

The aforementioned models only adopt segment-level labels for segment score calculation and model training. An extension~\cite{ye2018hybrid} proposed by Ye et al. demonstrates that incorporating word-level labels information can be beneficial for building semi-CRFs. The proposed Hybrid Semi-CRFs(HSCRF) model utilizes word-level and segment-level labels simultaneously to derive the segment scores. Besides, the methods of integrating CRF and HSCRF output layers into an unified network for jointly training and decoding are further presented.
The Hybrid Semi-CRFs model is also adopted as baseline for subsequent work
~\cite{liu2019towards}.

\paratitle{Skip-chain CRF}.
The Skip-chain CRF~\cite{sutton2012introduction} is a variant of conventional linear chain CRF that captures long-range label dependencies by means of skip edges, which basically refers to edges between the label positions not adjacent to each other. However, the skip-chain CRF contains loop in graph structure, making the process of model training and inference intractable. Loop belief propagation that requires multiple iterations of messaging can be one of the approximate solutions, but is fairly time consuming for large neural network based models. In order to mitigate the problem, Jagannatha et al.~\cite{jagannatha2016structured} propose an approximate approach for computation of marginals which adopts recurrent units to model the messages. The proposed approximate neural skip-chain CRF model is used for enhancing the exact phrase detection of clinical entities.

\paratitle{Embedded-State Latent CRF}.
Thai et al.~\cite{thai2018embedded} design a novel embedded-state latent CRF for neural sequence labeling, which has more capacities in modeling non-local label dependencies that often neglected by conventional CRF. This method incorporates latent variables into the CRF model for capturing global constraints between labels and applies representation learning to the output space. In order to reduce the numbers of parameters and prevent overfitting, a parsimonious factorized parameter strategy to learn low-rank embedding matrices are further adopted.

\paratitle{NCRF transducers}.
Based on the similar motivation of modeling long-range dependencies between labels, Hu et al.~\cite{hu2019neural} present a further extension and propose neural CRF transducers (NCRF transducers), which introduces RNN transducers to implement the edge potential in CRF model. The edge potential represents the score for current label by considering dependencies from all previous labels. Thus the proposed model can capture long-range label dependencies from the beginning up to each current position.

\subsubsection{Recurrent Neural Network}

\begin{figure}[!t]
	\centering
	\includegraphics[width=0.45\textwidth]{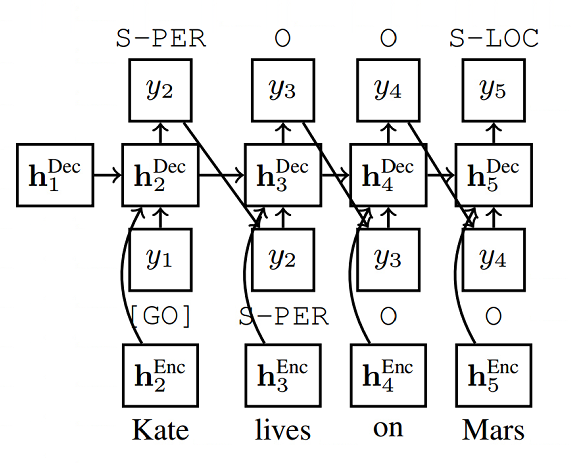}
	\caption{LSTM architecture for inferring label~\cite{shen2017deep}.}
	\label{fig:51}
\end{figure}

RNN is extremely suitable for feature extraction of sequential data, so it is widely used for encoding contextual information in sequence labeling models. Some studies demonstrate that RNN structure can also be adopted in the inference module for producing optimal label sequence. In addition to the learned representations output from context encoder, the information of former predicted labels also serves as an input. Thus the corresponding label of each word is generated based on both the features of input sequence and the previous predicted labels, making long-range label dependencies captured. However, unlike the global normalized CRF model, the RNN-based reasoning method greedily decodes the label from left to right, so it's a local normalized model that might suffer from label bias and exposure bias problems~\cite{andor2016globally}.

Shen et al.~\cite{shen2017deep} employ a LSTM layer on top of the context encoder for label decoding. As dipicted in Fig \ref{fig:51}, the decoder LSTM takes the last generated label as well as the contextual representation of current word as inputs, and computes the hidden state which will be passed through softmax function to finally decode the label. Zheng et al.~\cite{zheng2017joint} adopt a similar LSTM structure as the inference module of their sequence labeling model.

Unlike the above two studies, Vaswani et al.~\cite{vaswani2016supertagging} utilize a LSTM decoder that can be considered as parallel with the context encoder module. The LSTM only accepts the last label as input to produce a hidden state, which will be combined with the word context representation for label decoding. Zhang et al.~\cite{zhang2018learning} introduce a novel joint labeling strategy based on LSTM decoder. The output hidden state and contextual representation are not integrated before the labeling decision is made but independently estimate the labeling probability. Those two probabilities are then merged by weighted averaging to produce the final result. Specifically, a parameter is dynamically computed by a gate mechanism to adaptively balance the involvement of the two parts. Experiments show that the proposed label LSTM could significantly improve the performance.

\paratitle{Encoder-Decoder-Pointer Framework}.

\begin{figure}[!t]
	\centering
	\includegraphics[width=0.5\textwidth]{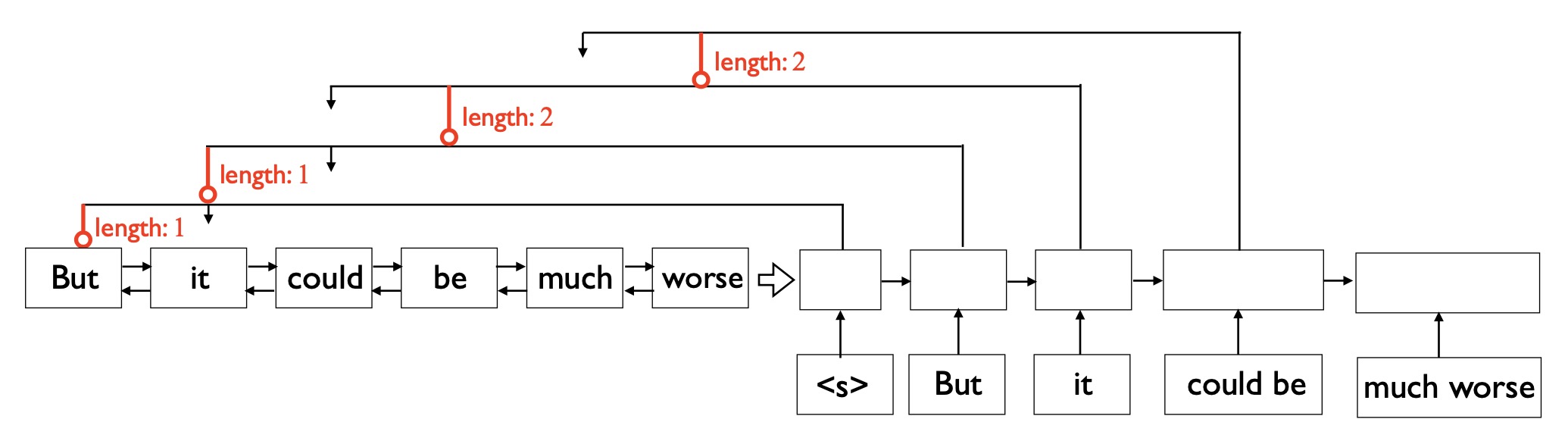}
	\caption{The encoder-decoder-pointer framework~\cite{zhai2017neural}.}
	\label{fig:44}
\end{figure}

Zhai et al.~\cite{zhai2017neural} propose a neural sequence chunking model based on an encoder-decoder-pointer framework, which is suitable for tasks that need assign labels to meaningful chunks in sentences, such as phrase chunking and semantic role labeling. The architecture is illustrated in Fig \ref{fig:44}. The proposed model divides original sequence labeling task into two steps: (1) Segmentation, identifying the scope of each chunk; (2) Labeling, treating each chunk as a complete unit to label. It adopts a pointer network~\cite{vinyals2015pointer} to process the segmentation by determining the ending point of each chunk and the LSTM decoder is utilized for labeling based on the segmentation results. The model proposed by Li et al.~\cite{li2018segbot} also employs the similar architecture for their text segmentation model, where a seq2seq model equipped with pointer network is designed to infer the segment boundaries.

\section{Evaluation Metrics and DataSets}
\label{datasets}
As mentioned in Section \ref{background}, three common related tasks of sequence labeling problems include POS tagging, NER, and chunking. In this section, we list some widely used datasets in Table \ref{table_datastes} and will describe several most commonly used datasets of these three tasks, and introduce the corresponding evaluation metrics as well.
\subsection{Datasets}

\begin{table*}[!t]
	\centering
	\caption{List of annotated datasets for POS and NER.}
    \setlength{\tabcolsep}{1pt}{
		\begin{tabular}{|c|c|c|c|}
			\hline
			\bf Task & \bf Corpus & \bf Year & \bf URL \\
			\hline
			{\multirow{5}{*}{POS}}
	~&Wall Street Journal(WSJ)& 2000 & \href{https://catalog.ldc.upenn.edu/LDC2000T43/}{https://catalog.ldc.upenn.edu/LDC2000T43/} \\
        \cline{2-4}
	~&NEGRA German Corpus& 2006 & \href{http://www.coli.uni-saarland.de/projects/sfb378/negra-corpus/}{http://www.coli.uni-saarland.de/projects/sfb378/negra-corpus/} \\
        \cline{2-4}
	~& Rit-Twitter & 2011 & \href{https://github.com/aritter/twitter\_nlp}{https://github.com/aritter/twitter\_nlp}   \\
	\cline{2-4}
	~& Prague Dependency Treebank  & 2012 - 2013 & \href{http://ufal.mff.cuni.cz/pdt2.0/}{http://ufal.mff.cuni.cz/pdt2.0/} \\
        \cline{2-4}
 	~& Universal Dependency(UD)  & 2015 - 2020 & \href{https://universaldependencies.org}{https://universaldependencies.org} \\
	\hline
    {\multirow{7}{*}{NER}}
     ~&ACE& 2000 - 2008 & \href{https://www.ldc.upenn.edu/collaborations/past-projects/ace}{https://www.ldc.upenn.edu/collaborations/past-projects/ace} \\
        \cline{2-4}
    ~& CoNLL02 & 2002 & \href{https://www.clips.uantwerpen.be/conll2002/ner/}{https://www.clips.uantwerpen.be/conll2002/ner/}  \\
	\cline{2-4}
	~& CoNLL03 & 2003 & \href{https://www.clips.uantwerpen.be/conll2003/ner/}{https://www.clips.uantwerpen.be/conll2003/ner/}  \\
	\cline{2-4}
	~& GENIA  & 2004 & \href{http://www.geniaproject.org/home}{http://www.geniaproject.org/home} \\
        \cline{2-4}
 	~& OntoNotes  & 2007 - 2012 & \href{https://catalog.ldc.upenn.edu/LDC2013T19}{https://catalog.ldc.upenn.edu/LDC2013T19} \\
 	\cline{2-4}
	~& WiNER  & 2012 & \href{http://rali.iro.umontreal.ca/rali/en/winer-wikipedia-for-ner}{http://rali.iro.umontreal.ca/rali/en/winer-wikipedia-for-ner} \\
        \cline{2-4}
 	~& W-NUT  & 2015 - 2018 & \href{http://noisy-text.github.io}{http://noisy-text.github.io} \\
	\hline
		\end{tabular}}
    \label{table_datastes}
\end{table*}

\subsubsection{POS tagging}
We will introduce three widely used datasets for part-of-speech tagging: WSJ, UD and Rit-Twitter.

\paratitle{WSJ}.
A standard dataset for POS tagging is the Wall Street Journal (WSJ) portion of the Penn Treebank~\cite{marcus1993building} and a large  number of work use it in their experiments. The dataset contains $25$ sections and classifies each word into $45$ different types of POS tags. A data split method used in ~\cite{collins2002discriminative} has become popular, in which sections $0$-$18$ as training data, $19$-$21$ as development data, and sections $22$-$24$ as test data.

\paratitle{UD}.
Universal Dependencies (UD) is a project that is developing cross-linguistic grammatical annotation, which contains more than $100$ treebanks in over $60$ languages. Its original annotation scheme for part-of-speech tagging take the form of Google universal POS tag sets~\cite{petrov2011universal} that include $12$ language-independent tags. A recent version of UD ~\cite{nivre2015universal} proposed a POS tag set that has 17 categories which partially overlap with those defined in~\cite{petrov2011universal}, and annotations from it have been used by many recent work~\cite{plank2016multilingual,berend2017sparse,Yasunaga2018Robust,kemos2018neural} to evaluate their models.

\paratitle{Rit-Twitter}.
The Rit-Twitter dataset~\cite{ritter2011named} is a benchmark for social media part-of-speech tagging which is comprised of $16K$ tokens from Twitter. It adopts an extended version of the PTB tagset with several Twitter-specific tags includes: retweets, @usernames, \#hashtags, and urls.

\subsubsection{NER}
We will introduce three widely used datasets for NER : CoNLL 2002, CoNLL 2003 and OntoNotes.

\paratitle{CoNLL 2002 $\&$ CoNLL 2003}.
CoNLL 2002~\cite{Sang2002Introduction} and CoNLL 2003~\cite{sang2003introduction} are two shared tasks created for NER. Both of these datasets contains annotations from newswire text and are tagged with four different entities - PER (person), LOC (location), ORG (organization) and MISC (miscellaneous including all other types of entities). CoNLL02 focuses on two languages: Dutch and Spanish, while CoNLL03 on English and German. Among them, the English dataset of CoNLL03 is the most widely used for NER and lots of recent work report their performance on it.

\paratitle{OntoNotes}.
The OntoNotes project~\cite{hovy2006ontonotes} was developed to annotate a large corpus from various genres in three languages (English, Chinese, and Arabic) with several layers of annotation, including named entities, coreference, part of speech, word sense, propositions, and syntactic parse trees. Regarding the NER dataset, the tag set consists of $18$ coarse entity types, containing $89$ subtypes and the whole dataset contains $2$ million tokens. There have been 5 versions so far, and the English dataset of the latest Release $5.0$ version~\cite{pradhan2013towards} has been utilized by many recent NER work in their experiments.

\subsubsection{Chunking}
\paratitle{CoNLL 2000}.
The CoNLL 2000 shared task~\cite{sang2000introduction} dataset is widely used for text chunking. The dataset is based on the WSJ part of the Penn Treebank as corpus and the annotation consists of $12$ different labels including $11$ syntactic chunk types in addition to Other. Since it only includes training and test sets, many researchers~\cite{Liu2017Empower, Peters2017Semi,Yang2016Transfer} randomly sampled a part of sentences from training set as the development set.

\subsection{Evaluation Metrics}
Part-of-speech tagging systems are usually evaluated according to the token accuracy. And F1-score, the harmonic mean of precision and recall, is usually adopted as the evaluation metric of NER and chunking.
\subsubsection{Accuracy}
Accuracy depicts the ratio of the number of correctly classified instances and the total number of instances, which can be computed using the following equation
$$
ACC = \frac{TP+TN}{TP+TN+FP+FN},
$$
where $TP,TN,FP,FN$ denote True positive, True negative, False positive, False negative, respectively.
\subsubsection{F1-score}
F1-score indicates the fraction of correctly classified instances for each class within the dataset, which can be computed as follows
$$
F1 = \frac{\sum_{i=1}^C{2*PERC_i*REC_i/(PREC_i+REC_i)}}{C},
$$
where $PREC$ is precision that is computed by $PREC = \frac{TP}{TP+FP}$, $REC$ is recall that can be computed by $REC = \frac{TP}{TP+FN}$, and $C$ denotes the total number of classes.

\section{Comparisons on Experimental Results of Various Techniques}
\label{results}

\begin{table}[!t]
    \hspace{-0.5cm}
    \caption{POS tagging accuracy of different models on test data from WSJ proportion of PTB.}
    \setlength{\tabcolsep}{1.5mm}{
		\begin{tabular}{|c|c|c|}
			\hline
			\bf External resources & \bf Method  & \bf Accuracy \\
			\hline
			{\multirow{22}{*}{None}}
	~&Collobert et al. 2011 ~\cite{Collobert2011Natural}&97.29\%\\
        \cline{2-3}
 	~&Santos et al. 2014~\cite{Santos2014Learning}&97.32\%\\
        \cline{2-3}
	~&Huang et al. 2015~\cite{huang2015bidirectional}&97.55\%\\
	\cline{2-3}
	~&Ling et al. 2015~\cite{ling2015finding}&97.78\%\\
        \cline{2-3}
 	~&Plank et al.2016~\cite{plank2016multilingual}&97.22\%\\ 
	\cline{2-3}
 	~&Rei et al. 2016~\cite{rei2016attending}&97.27\%\\
        \cline{2-3}
 	~&Vaswani et al. 2016  ~\cite{vaswani2016supertagging}&97.40\%\\
        \cline{2-3}
        ~&Andor et al. 2016~\cite{andor2016globally}&97.44\%\\
	\cline{2-3}
        &Ma and Hovy 2016~\cite{Ma2016End}&97.55\% \\
        \cline{2-3}
	 ~&Ma and Sun 2016~\cite{ma2016new}&97.56\%\\
	\cline{2-3}
        ~&Rei 2017~\cite{Rei2017Semi}&97.43\%\\
        \cline{2-3}     
        ~&Yang et al. 2017~\cite{Yang2016Transfer}&97.55\%\\
        \cline{2-3}
        ~&Kazi and Thompson 2017  ~\cite{kazi2017implicitly}&97.37\%\\
        \cline{2-3}
	~&Bohnet et al. 2018 ~\cite{bohnet2018morphosyntactic}&97.96\%\\
        \cline{2-3}
	~&Yasunaga et al. 2018~\cite{Yasunaga2018Robust}&97.55\%\\
        \cline{2-3}
        ~&Liu et al. 2018~\cite{Liu2017Empower}&97.53\%\\
        \cline{2-3}
        ~&Zhang et al. 2018~\cite{zhang2018learning}&97.59\%\\
        \cline{2-3}
        ~&Xin et al. 2018~\cite{xin2018learning}&97.58\%\\
        \cline{2-3}
        ~&Zhang et al. 2018~\cite{zhang2018sentence}&97.55\%\\
        \cline{2-3}
	~&Hu et al. 2019~\cite{hu2019neural}&97.52\%\\
    \cline{2-3}
        ~&Cui et al. 2019~\cite{cui2019hierarchically}&97.65\%\\
    \cline{2-3}
    ~&Jiang et al. 2020~\cite{jiang2019generalizing}&97.7\%\\
	\hline
    {\multirow{2}{*}{Unlabeled Word Corpus}}
        ~&Akbik et al. 2018  ~\cite{akbik2018contextual}&97.85\%\\
     \cline{2-3}
        ~&Clark et al. 2018  ~\cite{clark2018semi}&97.7\%\\
        \hline
		\end{tabular}}
    \label{table_pos}
\end{table}

\begin{table}[p]
    \hspace{-1.5cm}
    \caption{F1-score of different models on test data from CoNLL 2003 NER(English).}
	\begin{center}
    \setlength{\tabcolsep}{2mm}{
		\begin{tabular}{|c|c|c|}
			\hline
			\bf External resources & \bf Method  & \bf F1-score \\
			\hline
			{\multirow{25}{*}{None}}
	    ~&Collobert et al. 2011~\cite{Collobert2011Natural}&88.67\%\\
	    \cline{2-3}
            ~&Kuru et al. 2016~\cite{kuru2016charner}&84.52\%\\
            \cline{2-3}
 	    ~&Chiu and Nichols 2016~\cite{chiu2016named}&90.91\%\\
            \cline{2-3}
 	    ~&Lample et al. 2016~\cite{Lample2016Neural}&90.94\%\\
	    \cline{2-3}
            &Ma and Hovy 2016~\cite{Ma2016End}&91.21\% \\
            \cline{2-3}
	    ~&Rei 2017 ~\cite{Rei2017Semi}&86.26\%\\
            \cline{2-3}
            ~&Strubell et al. 2017 ~\cite{strubell2017fast}&90.54\%\\
            \cline{2-3}
	    ~&Zhang et al. 2017 ~\cite{Zhang2017Does}&90.70\%\\
            \cline{2-3}
  	    ~&Tran et al. 2017~\cite{tran2017named}&91.23\%\\
	    \cline{2-3}
	    ~&Wang et al. 2017  ~\cite{wang2017named}&91.24\%\\
            \cline{2-3}
           ~&Sato et al. 2017~\cite{sato2017segment}&91.28\%\\
            \cline{2-3}
 	   ~&Shen et al. 2018  ~\cite{shen2017deep}&90.69\%\\
            \cline{2-3}
	   ~&Zhang et al. 2018~\cite{zhang2018learning}&91.22\%\\
            \cline{2-3}
            ~& Liu et al. 2018 ~\cite{Liu2017Empower}&91.24\%\\
            \cline{2-3}
            ~&Ye and Ling 2018~\cite{ye2018hybrid}&91.38\%\\
            \cline{2-3}
            ~&Gregoric et al. 2018 ~\cite{gregoric2018named}&91.48\%\\ 
             ~&Zhang et al. 2018~\cite{zhang2018sentence}&91.57\%\\
            \cline{2-3}
            \cline{2-3}
            ~&Xin et al. 2018~\cite{xin2018learning}&91.64\%\\
            \cline{2-3}
            ~&Hu et al. 2019~\cite{hu2019neural}&91.40\%\\
            \cline{2-3}
            ~&Chen et al. 2019~\cite{chen2019grn}&91.44\%\\
            \cline{2-3}
             ~&Yan et al. 2019~\cite{yan2019tener}&91.45\%\\
            \cline{2-3}
             &Liu et al. 2019~\cite{liu2019gcdt}&91.96\% \\
              \cline{2-3}
               &Luo et al. 2020~\cite{luo2020hierarchical}&91.96\% \\
              \cline{2-3}
             &Jiang et al. 2020~\cite{jiang2019generalizing}&92.2\% \\
              \cline{2-3}
             &Li et al. 2020~\cite{li2020handling}&92.67\% \\
	    \hline
            CoNLL00,WSJ &Yang et al. 2017  ~\cite{Yang2016Transfer}&91.26\%\\
            \hline
            {\multirow{5}{*}{Gazetteers}}
             &Collobert et al. 2011~\cite{Collobert2011Natural}&89.59\% \\
             \cline{2-3}
            &Huang et al. 2015~\cite{huang2015bidirectional}&90.10\% \\
            \cline{2-3}
            &Wu et al. 2018~\cite{wu2018evaluating}&91.89\% \\
            \cline{2-3}
            &Liu et al. 2019~\cite{liu2019towards}&92.75\% \\
            \cline{2-3}
            &Chen et al. 2020~\cite{chen2020seqvat}&91.76\% \\
            \hline
            {\multirow{3}{*}{Lexicons}}
	    &Chiu and Nichols 2016~\cite{chiu2016named}&91.62\% \\
             \cline{2-3}
            &Sato et al. 2017~\cite{sato2017segment}&91.55\% \\
            \cline{2-3}
            &Ghaddar and Langlais 2018~\cite{ghaddar2018robust}&91.73\% \\
             \hline
            {\multirow{6}{*}{Unlabeled Word Corpus}}
             &Peters et al. 2017~\cite{Peters2017Semi}&91.93\% \\
             \cline{2-3}
            &Peters et al. 2018~\cite{Peters2018Deep}&92.22\% \\
            \cline{2-3}
            &Devlin et al. 2018~\cite{devlin2018bert}&92.80\% \\
            \cline{2-3}
            &Akbik et al. 2018~\cite{akbik2018contextual}&93.09\% \\
            \cline{2-3}
             &Clark et al. 2018~\cite{clark2018semi}&92.6\% \\
            \cline{2-3}
            &Li et al. 2020~\cite{li2019unified}&93.04\% \\
            \hline
            {\multirow{6}{*}{LM emb}}
             &Tran et al. 2017~\cite{tran2017named}&91.69\% \\
             \cline{2-3}
            &Chen et al. 2019~\cite{chen2019grn}&92.34\% \\
            \cline{2-3}
            &Hu et al. 2019~\cite{hu2019neural}&92.36\% \\
            \cline{2-3}
            &Liu et al. 2019~\cite{liu2019gcdt}&93.47\% \\
            \cline{2-3}
            &Jiang et al. 2019~\cite{jiang2019improved}&93.47\% \\
            \cline{2-3}
            &Luo et al. 2019~\cite{luo2020hierarchical}&93.37\% \\
             \hline
              Knowledge Graph &He et al. 2020  ~\cite{he2020knowledge-graph}&91.8\%\\
            \hline

		\end{tabular}}
	\end{center}
	
    \label{table_ner}
\end{table}

\begin{table}[h]
    \caption{F1-score of different models on test data from CoNLL 2000 Chunking.}
	\begin{center}
    \setlength{\tabcolsep}{1.5mm}{
		\begin{tabular}{|c|c|c|}
			\hline
			\bf External resources & \bf Method  & \bf F1-score \\
			\hline
			{\multirow{10}{*}{None}}
 	    ~&Collobert et al. 2011~\cite{Collobert2011Natural}&94.32\%\\
            \cline{2-3}
            ~&Huang et al. 2015~\cite{huang2015bidirectional}&94.46\%\\
            \cline{2-3}
	    ~&Rei et al. 2016~\cite{rei2016attending}&92.67\%\\
            \cline{2-3}	
            &Rei 2017~\cite{Rei2017Semi}&93.88\% \\
            \cline{2-3}
	    ~&Zhai et al. 2017~\cite{zhai2017neural}&94.72\%\\
            \cline{2-3}
            ~&Sato et al. 2017~\cite{sato2017segment}&94.84\%\\
	    \cline{2-3}
            ~&Zhang et al. 2017~\cite{Zhang2017Does}&95.01\%\\
            \cline{2-3}
            ~&Xin et al. 2018~\cite{xin2018learning}&95.29\%\\
            \cline{2-3}
            ~&Hu et al. 2019~\cite{hu2019neural}&95.14\%\\
            \cline{2-3}
            ~&Liu et al. 2019~\cite{liu2019gcdt}&95.43\%\\
            \cline{2-3}
             &Chen et al. 2020~\cite{chen2020seqvat}&95.45\% \\
            \hline
            {\multirow{3}{*}{Unlabeled Word Corpus}}
            ~&Peters et al. 2017~\cite{Peters2017Semi}&96.37\%\\
            \cline{2-3}
            ~&Akbik et al. 2018~\cite{akbik2018contextual}&96.72\%\\
             \cline{2-3}
            ~&Clark et al. 2018~\cite{clark2018semi}&97\%\\
             \hline
            LM emb&Liu et al. 2019~\cite{liu2019gcdt}&97.3\%\\
           \hline
            CoNLL03,WSJ& Yang et al. 2017~\cite{Yang2016Transfer}&95.41\%\\
            \hline
		\end{tabular}}
	\end{center}
    \label{table_chunking}
\end{table}

While formal experimental evaluation is left out of the scope of this paper, we present a brief analysis of the experimental results of various techniques. For each of these three tasks, we choose one widely used dataset and report the performance of various models on the benchmark. The three datasets includes WSJ for POS, CoNLL 2003 NER and CoNLL 2000 chunking, and the results for these three tasks are given in Table \ref{table_pos}, Table \ref{table_ner} and Table \ref{table_chunking}, respectively. We also indicate whether the model makes use of external knowledge or resource in these tables.

As shown in Table \ref{table_pos}, different models have achieved relatively high performance (more than $97\%$) in terms of the accuracy of POS tagging. Among these work listed in the table, the Bi-LSTM-CNN-CRF model proposed by Ma and Hovy~\cite{Ma2016End} has become a popular baseline for most subsequent work in this field, which is also the first end-to-end model for sequence labeling requiring no feature engineering or data preprocessing. The reported accuracy of Ma and Hovy is $97.55\%$, and several studies in recent two years slightly outperform it by exploring different issues and building new models. For example, the model proposed by Zhang et al.~\cite{zhang2018learning} performs better with a improvement of $0.04\%$, which takes the long range tag dependencies into consideration by incorporating a tag LSTM in their model. Besides, Bohnet et al.~\cite{bohnet2018morphosyntactic} achieves the state-of-the-art performance with $97.96\%$ accuracy by modeling the sentence-level context for initial character and word-based representations.

Table \ref{table_ner} shows the results of different models on CoNLL 2003 NER datatsets. Compared with the POS tagging task, the overall score of NER task is lower, with most work between $91\%$ and $92\%$, which indicates NER is more difficult than POS tagging.
 Among the work that utilize no external resources, Li et al.~\cite{li2020handling} performs best, with a average F1-score of $92.67\%$. 
 Their proposed model focuses on rare entities and applied novel techniques including 
local context reconstruction and delexicalized entity identification.
 We can observe that models which utilize external resources can generally achieve higher performance on all these three tasks, especially pretraining language models that using large unlabeled word corpus. But these models require a larger neural network that need huge computing resources and longer time for training.

\section{The Promising Paths for Future Research}
\label{future}
Although much success has been achieved in this filed, challenges still exist from different perspectives. In this section, we provide the following directions for further research in deep learning based sequence labeling.

\paratitle{Sequence labeling for low-resource data.}
Supervised learning algorithms including deep learning based models, rely on large annotated data for training. However, data annotations are expensive and often take a lot of time, leaving a big challenge in sequence labeling for many low-resource languages and specific resource-poor domains. Although some work have explored methods for this problem, there still exists a large scope for improvement. Future efforts could be dedicated on enhancing performance of sequence labeling on low-resource data by focusing on the following three research directions: (1) training a LM like BERT with the unlabeled corpus and finetune it with limited labeled corpus in a low-resource data; (2) providing more effective deep transfer learning models to transfer knowledge from one language or domain to another; (3) exploring appropriate data augmentation techniques to enlarge the available data for sequence labeling.

\paratitle{Scalability of deep learning based sequence labeling.}
Most neural models for sequence labeling do not scale well for large data, making it a challenge to build more scalable deep learning based sequence labeling models. The main reason for this is when the size of data grows, the parameters of models increase exponentially, leading to the high complexity of back propagation. While several models have achieved excellent performance with huge computing power, there exists need for developing approaches to balance model complexity and scalability. In addition, for pratical usage, its necessary to develop scalable methods for real-world applications.

\paratitle{Utilization of external resources.}
As discussed in Section \ref{results}, the performance of neural sequence labeling models benefits significantly from external resources, including gazetteers, lexicons, large unlabeled word corpus, and etc. Though some research effort have been dedicated on this issue, how to effectively incorporate external resources in neural sequence labeling models remains to be explored.

\section{Conclusions}
\label{conclusion}

This survey aims to thoroughly review applications of deep learning techniques in sequence labeling, and provides a panoramic view so that readers can build a comprehensive understanding of this area. We present a summary for the literature with a scientific taxonomy. In addition, we provide an overview of the datasets and evaluation metrics of the commonly studied tasks of sequence labeling problems. Besides, we also discuss and compare the results of different models and analyze the factors and different architectures that affect the performance. Finally, we present readers with the challenges and open issues faced by current methods and identify the future directions in this area. We hope that this survey can help to enlighten and guide the researchers, practitioners, and educators who are interested in sequence labeling.

\bibliography{main}

\begin{thebibliography}{100}

\bibitem{akbik2018contextual}
Alan Akbik, Duncan Blythe, and Roland Vollgraf.
\newblock Contextual string embeddings for sequence labeling.
\newblock In {\em COLING}, pages 1638--1649, 2018.

\bibitem{andor2016globally}
Daniel Andor, Chris Alberti, David Weiss, Aliaksei Severyn, Alessandro Presta,
  Kuzman Ganchev, Slav Petrov, and Michael Collins.
\newblock Globally normalized transition-based neural networks.
\newblock {\em arXiv preprint arXiv:1603.06042}, 2016.

\bibitem{baum1966statistical}
Leonard~E Baum and Ted Petrie.
\newblock Statistical inference for probabilistic functions of finite state
  markov chains.
\newblock {\em The annals of mathematical statistics}, 37(6):1554--1563, 1966.

\bibitem{bekoulis2019sub}
Giannis Bekoulis, Johannes Deleu, Thomas Demeester, and Chris Develder.
\newblock Sub-event detection from twitter streams as a sequence labeling
  problem.
\newblock {\em NAACL}, 2019.

\bibitem{bender2003maximum}
Oliver Bender, Franz~Josef Och, and Hermann Ney.
\newblock Maximum entropy models for named entity recognition.
\newblock In {\em Proceedings of the seventh conference on Natural language
  learning at HLT-NAACL 2003-Volume 4}, pages 148--151. Association for
  Computational Linguistics, 2003.

\bibitem{berend2017sparse}
G{\'a}abor Berend.
\newblock Sparse coding of neural word embeddings for multilingual sequence
  labeling.
\newblock {\em Transactions of the Association for Computational Linguistics},
  5:247--261, 2017.

\bibitem{bikel1999algorithm}
Daniel~M Bikel, Richard Schwartz, and Ralph~M Weischedel.
\newblock An algorithm that learns what's in a name.
\newblock {\em Machine learning}, 34(1-3):211--231, 1999.

\bibitem{bohnet2018morphosyntactic}
Bernd Bohnet, Ryan McDonald, Goncalo Simoes, Daniel Andor, Emily Pitler, and
  Joshua Maynez.
\newblock Morphosyntactic tagging with a meta-bilstm model over context
  sensitive token encodings.
\newblock {\em ACL}, 2018.

\bibitem{cao2018adversarial}
Pengfei Cao, Yubo Chen, Kang Liu, Jun Zhao, and Shengping Liu.
\newblock Adversarial transfer learning for chinese named entity recognition
  with self-attention mechanism.
\newblock In {\em Proceedings of the 2018 Conference on Empirical Methods in
  Natural Language Processing}, pages 182--192, 2018.

\bibitem{chen2019grn}
Hui Chen, Zijia Lin, Guiguang Ding, Jianguang Lou, Yusen Zhang, and Borje
  Karlsson.
\newblock Grn: Gated relation network to enhance convolutional neural network
  for named entity recognition.
\newblock {\em AAAI}, 2019.

\bibitem{chen2020seqvat}
Luoxin Chen, Weitong Ruan, Xinyue Liu, and Jianhua Lu.
\newblock Seqvat: Virtual adversarial training for semi-supervised sequence
  labeling.
\newblock In {\em Proceedings of the 58th Annual Meeting of the Association for
  Computational Linguistics}, pages 8801--8811, 2020.

\bibitem{chieu2002named}
Hai~Leong Chieu and Hwee~Tou Ng.
\newblock Named entity recognition: a maximum entropy approach using global
  information.
\newblock In {\em Proceedings of the 19th international conference on
  Computational linguistics-Volume 1}, pages 1--7. Association for
  Computational Linguistics, 2002.

\bibitem{chiu2016named}
Jason~PC Chiu and Eric Nichols.
\newblock Named entity recognition with bidirectional lstm-cnns.
\newblock {\em Transactions of the Association for Computational Linguistics},
  4:357--370, 2016.

\bibitem{cho2014properties}
Kyunghyun Cho, Bart Van~Merri{\"e}nboer, Dzmitry Bahdanau, and Yoshua Bengio.
\newblock On the properties of neural machine translation: Encoder-decoder
  approaches.
\newblock {\em arXiv preprint arXiv:1409.1259}, 2014.

\bibitem{clark2018semi}
Kevin Clark, Minh-Thang Luong, Christopher~D Manning, and Quoc~V Le.
\newblock Semi-supervised sequence modeling with cross-view training.
\newblock {\em EMNLP}, 2018.

\bibitem{collins2002discriminative}
Michael Collins.
\newblock Discriminative training methods for hidden markov models: Theory and
  experiments with perceptron algorithms.
\newblock In {\em Proceedings of the ACL-02 conference on Empirical methods in
  natural language processing-Volume 10}, pages 1--8. Association for
  Computational Linguistics, 2002.

\bibitem{Collobert2011Natural}
Ronan Collobert, Koray Kavukcuoglu, Jason Weston, Leon Bottou, Pavel Kuksa, and
  Michael Karlen.
\newblock Natural language processing (almost) from scratch.
\newblock {\em Journal of Machine Learning Research}, 12(1):2493--2537, 2011.

\bibitem{cui2019hierarchically}
Leyang Cui and Yue Zhang.
\newblock Hierarchically-refined label attention network for sequence labeling.
\newblock {\em EMNLP}, 2019.

\bibitem{devlin2018bert}
Jacob Devlin, Ming-Wei Chang, Kenton Lee, and Kristina Toutanova.
\newblock Bert: Pre-training of deep bidirectional transformers for language
  understanding.
\newblock {\em arXiv preprint arXiv:1810.04805}, 2018.

\bibitem{dozat2017stanford}
Timothy Dozat, Peng Qi, and Christopher~D Manning.
\newblock Stanford's graph-based neural dependency parser at the conll 2017
  shared task.
\newblock In {\em Proceedings of the CoNLL 2017 Shared Task: Multilingual
  Parsing from Raw Text to Universal Dependencies}, pages 20--30, 2017.

\bibitem{eddy1996hidden}
Sean~R Eddy.
\newblock Hidden markov models.
\newblock {\em Current opinion in structural biology}, 6(3):361--365, 1996.

\bibitem{feng2018improving}
Xiaocheng Feng, Xiachong Feng, Bing Qin, Zhangyin Feng, and Ting Liu.
\newblock Improving low resource named entity recognition using cross-lingual
  knowledge transfer.
\newblock In {\em IJCAI}, pages 4071--4077, 2018.

\bibitem{ghaddar2018robust}
Abbas Ghaddar and Philippe Langlais.
\newblock Robust lexical features for improved neural network named-entity
  recognition.
\newblock {\em arXiv preprint arXiv:1806.03489}, 2018.

\bibitem{G2018Constituent}
Carlos G{\'o}mez-Rodr{\'\i}guez and David Vilares.
\newblock Constituent parsing as sequence labeling.
\newblock {\em EMNLP}, 2018.

\bibitem{gooding2019complex}
Sian Gooding and Ekaterina Kochmar.
\newblock Complex word identification as a sequence labelling task.
\newblock {\em ACL}, 2019.

\bibitem{gregoric2018named}
Andrej~Zukov Gregoric, Yoram Bachrach, and Sam Coope.
\newblock Named entity recognition with parallel recurrent neural networks.
\newblock In {\em Proceedings of the 56th Annual Meeting of the Association for
  Computational Linguistics (Volume 2: Short Papers)}, pages 69--74, 2018.

\bibitem{gui2017part}
Tao Gui, Qi~Zhang, Haoran Huang, Minlong Peng, and Xuanjing Huang.
\newblock Part-of-speech tagging for twitter with adversarial neural networks.
\newblock In {\em Proceedings of the 2017 Conference on Empirical Methods in
  Natural Language Processing}, pages 2411--2420, 2017.

\bibitem{guo2019star}
Qipeng Guo, Xipeng Qiu, Pengfei Liu, Yunfan Shao, Xiangyang Xue, and Zheng
  Zhang.
\newblock Star-transformer.
\newblock {\em NAACL}, 2019.

\bibitem{hammerton2003named}
James Hammerton.
\newblock Named entity recognition with long short-term memory.
\newblock In {\em Proceedings of the seventh conference on Natural language
  learning at HLT-NAACL 2003-Volume 4}, pages 172--175. Association for
  Computational Linguistics, 2003.

\bibitem{he2016deep}
Kaiming He, Xiangyu Zhang, Shaoqing Ren, and Jian Sun.
\newblock Deep residual learning for image recognition.
\newblock In {\em Proceedings of the IEEE conference on computer vision and
  pattern recognition}, pages 770--778, 2016.

\bibitem{he2020knowledge-graph}
Qizhen He, Liang Wu, Yida Yin, and Heming Cai.
\newblock Knowledge-graph augmented word representations for named entity
  recognition.
\newblock {\em AAAI}, 2020.

\bibitem{hearst1998support}
Marti~A. Hearst, Susan~T Dumais, Edgar Osuna, John Platt, and Bernhard
  Scholkopf.
\newblock Support vector machines.
\newblock {\em IEEE Intelligent Systems and their applications}, 13(4):18--28,
  1998.

\bibitem{hovy2006ontonotes}
Eduard Hovy, Mitchell Marcus, Martha Palmer, Lance Ramshaw, and Ralph
  Weischedel.
\newblock Ontonotes: The 90$\backslash$\% solution.
\newblock In {\em Proceedings of the human language technology conference of
  the NAACL, Companion Volume: Short Papers}, 2006.

\bibitem{hu2019neural}
Kai Hu, Zhijian Ou, Min Hu, and Junlan Feng.
\newblock Neural crf transducers for sequence labeling.
\newblock In {\em ICASSP 2019-2019 IEEE International Conference on Acoustics,
  Speech and Signal Processing (ICASSP)}, pages 2997--3001. IEEE, 2019.

\bibitem{huang2015bidirectional}
Zhiheng Huang, Wei Xu, and Kai Yu.
\newblock Bidirectional lstm-crf models for sequence tagging.
\newblock {\em Computer Science}, 2015.

\bibitem{isozaki2002efficient}
Hideki Isozaki and Hideto Kazawa.
\newblock Efficient support vector classifiers for named entity recognition.
\newblock In {\em Proceedings of the 19th international conference on
  Computational linguistics-Volume 1}, pages 1--7. Association for
  Computational Linguistics, 2002.

\bibitem{jagannatha2016structured}
Abhyuday~N Jagannatha and Hong Yu.
\newblock Structured prediction models for rnn based sequence labeling in
  clinical text.
\newblock In {\em Proceedings of the conference on empirical methods in natural
  language processing. conference on empirical methods in natural language
  processing}, volume 2016, page 856. NIH Public Access, 2016.

\bibitem{jiang2019improved}
Yufan Jiang, Chi Hu, Tong Xiao, Chunliang Zhang, and Jingbo Zhu.
\newblock Improved differentiable architecture search for language modeling and
  named entity recognition.
\newblock In {\em Proceedings of the 2019 Conference on Empirical Methods in
  Natural Language Processing and the 9th International Joint Conference on
  Natural Language Processing (EMNLP-IJCNLP)}, pages 3576--3581, 2019.

\bibitem{jiang2019generalizing}
Zhengbao Jiang, Wei Xu, Jun Araki, and Graham Neubig.
\newblock Generalizing natural language analysis through span-relation
  representations.
\newblock {\em ACL}, 2020.

\bibitem{kann2018character}
Katharina Kann, Johannes Bjerva, Isabelle Augenstein, Barbara Plank, and Anders
  S{\o}gaard.
\newblock Character-level supervision for low-resource pos tagging.
\newblock In {\em Proceedings of the Workshop on Deep Learning Approaches for
  Low-Resource NLP}, pages 1--11, 2018.

\bibitem{kapur1989maximum}
Jagat~Narain Kapur.
\newblock {\em Maximum-entropy models in science and engineering}.
\newblock John Wiley \& Sons, 1989.

\bibitem{kazi2017implicitly}
Michaeel Kazi and Brian Thompson.
\newblock Implicitly-defined neural networks for sequence labeling.
\newblock In {\em Proceedings of the 55th Annual Meeting of the Association for
  Computational Linguistics (Volume 2: Short Papers)}, pages 172--177, 2017.

\bibitem{kemos2018neural}
Apostolos Kemos, Heike Adel, and Hinrich Sch{\"u}tze.
\newblock Neural semi-markov conditional random fields for robust
  character-based part-of-speech tagging.
\newblock {\em arXiv preprint arXiv:1808.04208}, 2018.

\bibitem{kim2017cross}
Joo-Kyung Kim, Young-Bum Kim, Ruhi Sarikaya, and Eric Fosler-Lussier.
\newblock Cross-lingual transfer learning for pos tagging without cross-lingual
  resources.
\newblock In {\em Proceedings of the 2017 Conference on Empirical Methods in
  Natural Language Processing}, pages 2832--2838, 2017.

\bibitem{knoll2004jacobian}
Dana~A Knoll and David~E Keyes.
\newblock Jacobian-free newton--krylov methods: a survey of approaches and
  applications.
\newblock {\em Journal of Computational Physics}, 193(2):357--397, 2004.

\bibitem{kong2015segmental}
Lingpeng Kong, Chris Dyer, and Noah~A Smith.
\newblock Segmental recurrent neural networks.
\newblock {\em arXiv preprint arXiv:1511.06018}, 2015.

\bibitem{krishnan2006effective}
Vijay Krishnan and Christopher~D Manning.
\newblock An effective two-stage model for exploiting non-local dependencies in
  named entity recognition.
\newblock In {\em Proceedings of the 21st International Conference on
  Computational Linguistics and the 44th annual meeting of the Association for
  Computational Linguistics}, pages 1121--1128. Association for Computational
  Linguistics, 2006.

\bibitem{kudoh2000use}
Taku Kudoh and Yuji Matsumoto.
\newblock Use of support vector learning for chunk identification.
\newblock In {\em Fourth Conference on Computational Natural Language Learning
  and the Second Learning Language in Logic Workshop}, 2000.

\bibitem{kumar2010part}
Dinesh Kumar and Gurpreet~Singh Josan.
\newblock Part of speech taggers for morphologically rich indian languages: a
  survey.
\newblock {\em International Journal of Computer Applications}, 6(5):32--41,
  2010.

\bibitem{kuru2016charner}
Onur Kuru, Ozan~Arkan Can, and Deniz Yuret.
\newblock Charner: Character-level named entity recognition.
\newblock In {\em Proceedings of COLING 2016, the 26th International Conference
  on Computational Linguistics: Technical Papers}, pages 911--921, 2016.

\bibitem{lafferty2001conditional}
John Lafferty, Andrew McCallum, and Fernando~CN Pereira.
\newblock Conditional random fields: Probabilistic models for segmenting and
  labeling sequence data.
\newblock In {\em ICML}, 2001.

\bibitem{Lample2016Neural}
Guillaume Lample, Miguel Ballesteros, Sandeep Subramanian, Kazuya Kawakami, and
  Chris Dyer.
\newblock Neural architectures for named entity recognition.
\newblock In {\em NAACL}, 2016.

\bibitem{lei2016layer}
Jimmy Lei~Ba, Jamie~Ryan Kiros, and Geoffrey~E Hinton.
\newblock Layer normalization.
\newblock {\em arXiv preprint arXiv:1607.06450}, 2016.

\bibitem{li2018survey}
Jing Li, Aixin Sun, Jianglei Han, and Chenliang Li.
\newblock A survey on deep learning for named entity recognition.
\newblock {\em arXiv preprint arXiv:1812.09449}, 2018.

\bibitem{li2018segbot}
Jing Li, Aixin Sun, and Shafiq Joty.
\newblock Segbot: A generic neural text segmentation model with pointer
  network.
\newblock In {\em IJCAI}, pages 4166--4172, 2018.

\bibitem{li2016dataset}
Peng Li, Wei Li, Zhengyan He, Xuguang Wang, Ying Cao, Jie Zhou, and Wei Xu.
\newblock Dataset and neural recurrent sequence labeling model for open-domain
  factoid question answering.
\newblock {\em arXiv preprint arXiv:1607.06275}, 2016.

\bibitem{li2019unified}
Xiaoya Li, Jingrong Feng, Yuxian Meng, Qinghong Han, Fei Wu, and Jiwei Li.
\newblock A unified mrc framework for named entity recognition.
\newblock {\em ACL}, 2020.

\bibitem{li2020handling}
Yangming Li, Han Li, Kaisheng Yao, and Xiaolong Li.
\newblock Handling rare entities for neural sequence labeling.
\newblock In {\em Proceedings of the 58th Annual Meeting of the Association for
  Computational Linguistics}, pages 6441--6451, 2020.

\bibitem{li2004svm}
Yaoyong Li, Kalina Bontcheva, and Hamish Cunningham.
\newblock Svm based learning system for information extraction.
\newblock In {\em International Workshop on Deterministic and Statistical
  Methods in Machine Learning}, pages 319--339. Springer, 2004.

\bibitem{li2018seq2seq}
Zuchao Li, Jiaxun Cai, Shexia He, and Hai Zhao.
\newblock Seq2seq dependency parsing.
\newblock In {\em Proceedings of the 27th International Conference on
  Computational Linguistics}, pages 3203--3214, 2018.

\bibitem{lin2019gazetteer}
Hongyu Lin, Yaojie Lu, Xianpei Han, Le~Sun, Bin Dong, and Shanshan Jiang.
\newblock Gazetteer-enhanced attentive neural networks for named entity
  recognition.
\newblock In {\em Proceedings of the 2019 Conference on Empirical Methods in
  Natural Language Processing and the 9th International Joint Conference on
  Natural Language Processing (EMNLP-IJCNLP)}, pages 6233--6238, 2019.

\bibitem{ling2015finding}
Wang Ling, Tiago Lu{\'\i}s, Lu{\'\i}s Marujo, Ram{\'o}n~Fernandez Astudillo,
  Silvio Amir, Chris Dyer, Alan~W Black, and Isabel Trancoso.
\newblock Finding function in form: Compositional character models for open
  vocabulary word representation.
\newblock {\em arXiv preprint arXiv:1508.02096}, 2015.

\bibitem{ling2015not}
Wang Ling, Yulia Tsvetkov, Silvio Amir, Ramon Fermandez, Chris Dyer, Alan~W
  Black, Isabel Trancoso, and Chu-Cheng Lin.
\newblock Not all contexts are created equal: Better word representations with
  variable attention.
\newblock In {\em Proceedings of the 2015 Conference on Empirical Methods in
  Natural Language Processing}, pages 1367--1372, 2015.

\bibitem{Liu2017Heterogeneous}
Liyuan Liu, Xiang Ren, Qi~Zhu, Shi Zhi, Huan Gui, Heng Ji, and Jiawei Han.
\newblock Heterogeneous supervision for relation extraction: A representation
  learning approach.
\newblock In {\em EMNLP}, 2017.

\bibitem{Liu2017Empower}
Liyuan Liu, Jingbo Shang, Xiang Ren, Frank~Fangzheng Xu, Huan Gui, Jian Peng,
  and Jiawei Han.
\newblock Empower sequence labeling with task-aware neural language model.
\newblock In {\em AAAI}, 2018.

\bibitem{liu2019towards}
Tianyu Liu, Jin-Ge Yao, and Chin-Yew Lin.
\newblock Towards improving neural named entity recognition with gazetteers.
\newblock In {\em Proceedings of the 57th Annual Meeting of the Association for
  Computational Linguistics}, pages 5301--5307, 2019.

\bibitem{liu2019gcdt}
Yijin Liu, Fandong Meng, Jinchao Zhang, Jinan Xu, Yufeng Chen, and Jie Zhou.
\newblock Gcdt: A global context enhanced deep transition architecture for
  sequence labeling.
\newblock {\em ACL}, 2019.

\bibitem{luo2020hierarchical}
Ying Luo, Fengshun Xiao, and Hai Zhao.
\newblock Hierarchical contextualized representation for named entity
  recognition.
\newblock In {\em AAAI}, pages 8441--8448, 2020.

\bibitem{ma2017jointly}
Mingbo Ma, Kai Zhao, Liang Huang, Bing Xiang, and Bowen Zhou.
\newblock Jointly trained sequential labeling and classification by sparse
  attention neural networks.
\newblock {\em arXiv preprint arXiv:1709.10191}, 2017.

\bibitem{ma2016new}
Shuming Ma and Xu~Sun.
\newblock A new recurrent neural crf for learning non-linear edge features.
\newblock {\em arXiv preprint arXiv:1611.04233}, 2016.

\bibitem{Ma2016End}
Xuezhe Ma and Eduard Hovy.
\newblock End-to-end sequence labeling via bi-directional lstm-cnns-crf.
\newblock In {\em ACL}, 2016.

\bibitem{marcus1993building}
Mitchell Marcus, Beatrice Santorini, and Mary~Ann Marcinkiewicz.
\newblock Building a large annotated corpus of english: The penn treebank.
\newblock 1993.

\bibitem{mccallum2000maximum}
Andrew McCallum, Dayne Freitag, and Fernando~CN Pereira.
\newblock Maximum entropy markov models for information extraction and
  segmentation.
\newblock In {\em Icml}, volume~17, pages 591--598, 2000.

\bibitem{mccallum2003early}
Andrew McCallum and Wei Li.
\newblock Early results for named entity recognition with conditional random
  fields, feature induction and web-enhanced lexicons.
\newblock In {\em Proceedings of the seventh conference on Natural language
  learning at HLT-NAACL 2003-Volume 4}, pages 188--191. Association for
  Computational Linguistics, 2003.

\bibitem{mikolov2013efficient}
Tomas Mikolov, Kai Chen, Greg Corrado, and Jeffrey Dean.
\newblock Efficient estimation of word representations in vector space.
\newblock {\em arXiv preprint arXiv:1301.3781}, 2013.

\bibitem{nadeau2007survey}
David Nadeau and Satoshi Sekine.
\newblock A survey of named entity recognition and classification.
\newblock {\em Lingvisticae Investigationes}, 30(1):3--26, 2007.

\bibitem{nallapati2017summarunner}
Ramesh Nallapati, Feifei Zhai, and Bowen Zhou.
\newblock Summarunner: A recurrent neural network based sequence model for
  extractive summarization of documents.
\newblock In {\em Thirty-First AAAI Conference on Artificial Intelligence},
  2017.

\bibitem{Ng2010Supervised}
V.~Ng.
\newblock Supervised noun phrase coreference research: The first fifteen years.
\newblock In {\em ACL}, 2010.

\bibitem{nguyen2007comparisons}
Nam Nguyen and Yunsong Guo.
\newblock Comparisons of sequence labeling algorithms and extensions.
\newblock In {\em Proceedings of the 24th international conference on Machine
  learning}, pages 681--688. ACM, 2007.

\bibitem{nivre2015universal}
Joakim Nivre, {\v{Z}}eljko Agi{\'c}, Maria~Jesus Aranzabe, Masayuki Asahara,
  Aitziber Atutxa, Miguel Ballesteros, John Bauer, Kepa Bengoetxea, Riyaz~Ahmad
  Bhat, Cristina Bosco, et~al.
\newblock Universal dependencies 1.2.
\newblock 2015.

\bibitem{Nivre2004Deterministic}
Joakim Nivre and Mario Scholz.
\newblock Deterministic dependency parsing of english text.
\newblock In {\em COLING}, page~64. Association for Computational Linguistics,
  2004.

\bibitem{park2019selectively}
Jaehui Park.
\newblock Selectively connected self-attentions for semantic role labeling.
\newblock {\em Applied Sciences}, 9(8):1716, 2019.

\bibitem{pascanu2013construct}
Razvan Pascanu, Caglar Gulcehre, Kyunghyun Cho, and Yoshua Bengio.
\newblock How to construct deep recurrent neural networks.
\newblock {\em ICLR}, 2014.

\bibitem{pennington2014glove}
Jeffrey Pennington, Richard Socher, and Christopher Manning.
\newblock Glove: Global vectors for word representation.
\newblock In {\em Proceedings of the 2014 conference on empirical methods in
  natural language processing (EMNLP)}, pages 1532--1543, 2014.

\bibitem{Peters2017Semi}
Matthew~E Peters, Waleed Ammar, Chandra Bhagavatula, and Russell Power.
\newblock Semi-supervised sequence tagging with bidirectional language models.
\newblock In {\em ACL}, 2017.

\bibitem{Peters2018Deep}
Matthew~E. Peters, Mark Neumann, Mohit Iyyer, Matt Gardner, Christopher Clark,
  Kenton Lee, and Luke Zettlemoyer.
\newblock Deep contextualized word representations.
\newblock {\em NAACL}, 2018.

\bibitem{petrov2011universal}
Slav Petrov, Dipanjan Das, and Ryan McDonald.
\newblock A universal part-of-speech tagset.
\newblock {\em arXiv preprint arXiv:1104.2086}, 2011.

\bibitem{plank2016multilingual}
Barbara Plank, Anders S{\o}gaard, and Yoav Goldberg.
\newblock Multilingual part-of-speech tagging with bidirectional long
  short-term memory models and auxiliary loss.
\newblock {\em ACL}, 2016.

\bibitem{pradhan2013towards}
Sameer Pradhan, Alessandro Moschitti, Nianwen Xue, Hwee~Tou Ng, Anders
  Bj{\"o}rkelund, Olga Uryupina, Yuchen Zhang, and Zhi Zhong.
\newblock Towards robust linguistic analysis using ontonotes.
\newblock In {\em Proceedings of the Seventeenth Conference on Computational
  Natural Language Learning}, pages 143--152, 2013.

\bibitem{ratnaparkhi1996maximum}
Adwait Ratnaparkhi.
\newblock A maximum entropy model for part-of-speech tagging.
\newblock In {\em Conference on Empirical Methods in Natural Language
  Processing}, 1996.

\bibitem{Rei2017Semi}
Marek Rei.
\newblock Semi-supervised multitask learning for sequence labeling.
\newblock In {\em ACL}, 2017.

\bibitem{rei2016attending}
Marek Rei, Gamal~KO Crichton, and Sampo Pyysalo.
\newblock Attending to characters in neural sequence labeling models.
\newblock {\em COLING}, 2016.

\bibitem{rei2016compositional}
Marek Rei and Helen Yannakoudakis.
\newblock Compositional sequence labeling models for error detection in learner
  writing.
\newblock {\em arXiv preprint arXiv:1607.06153}, 2016.

\bibitem{rijhwani2020soft}
Shruti Rijhwani, Shuyan Zhou, Graham Neubig, and Jaime Carbonell.
\newblock Soft gazetteers for low-resource named entity recognition.
\newblock {\em ACL}, 2020.

\bibitem{ritter2011named}
Alan Ritter, Sam Clark, Oren Etzioni, et~al.
\newblock Named entity recognition in tweets: an experimental study.
\newblock In {\em Proceedings of the conference on empirical methods in natural
  language processing}, pages 1524--1534. Association for Computational
  Linguistics, 2011.

\bibitem{sang2000introduction}
Erik~F Sang and Sabine Buchholz.
\newblock Introduction to the conll-2000 shared task: Chunking.
\newblock {\em arXiv preprint cs/0009008}, 2000.

\bibitem{sang2003introduction}
Erik~F Sang and Fien De~Meulder.
\newblock Introduction to the conll-2003 shared task: Language-independent
  named entity recognition.
\newblock {\em arXiv preprint cs/0306050}, 2003.

\bibitem{Sang2002Introduction}
Erik F. Tjong~Kim Sang.
\newblock Introduction to the conll-2002 shared task: Language-independent
  named entity recognition.
\newblock {\em Computer Science}, pages 142--147, 2002.

\bibitem{Santos2014Learning}
Cicero Nogueira~Dos Santos and Bianca Zadrozny.
\newblock Learning character-level representations for part-of-speech tagging.
\newblock In {\em ICML}, 2014.

\bibitem{sarawagi2005semi}
Sunita Sarawagi and William~W Cohen.
\newblock Semi-markov conditional random fields for information extraction.
\newblock In {\em Advances in neural information processing systems}, pages
  1185--1192, 2005.

\bibitem{sato2017segment}
Motoki Sato, Hiroyuki Shindo, Ikuya Yamada, and Yuji Matsumoto.
\newblock Segment-level neural conditional random fields for named entity
  recognition.
\newblock In {\em Proceedings of the Eighth International Joint Conference on
  Natural Language Processing (Volume 2: Short Papers)}, pages 97--102, 2017.

\bibitem{saxena2018emotionx}
Rohit Saxena, Savita Bhat, and Niranjan Pedanekar.
\newblock Emotionx-area66: Predicting emotions in dialogues using hierarchical
  attention network with sequence labeling.
\newblock In {\em Proceedings of the Sixth International Workshop on Natural
  Language Processing for Social Media}, pages 50--55, 2018.

\bibitem{shen2017deep}
Yanyao Shen, Hyokun Yun, Zachary~C Lipton, Yakov Kronrod, and Animashree
  Anandkumar.
\newblock Deep active learning for named entity recognition.
\newblock {\em ICLR}, 2018.

\bibitem{strubell2017fast}
Emma Strubell, Patrick Verga, David Belanger, and Andrew McCallum.
\newblock Fast and accurate entity recognition with iterated dilated
  convolutions.
\newblock {\em EMNLP}, 2017.

\bibitem{strzyz2019viable}
Michalina Strzyz, David Vilares, and Carlos G{\'o}mez-Rodr{\'\i}guez.
\newblock Viable dependency parsing as sequence labeling.
\newblock {\em NAACL}, 2019.

\bibitem{sutton2012introduction}
Charles Sutton, Andrew McCallum, et~al.
\newblock An introduction to conditional random fields.
\newblock {\em Foundations and Trends{\textregistered} in Machine Learning},
  4(4):267--373, 2012.

\bibitem{tan2018deep}
Zhixing Tan, Mingxuan Wang, Jun Xie, Yidong Chen, and Xiaodong Shi.
\newblock Deep semantic role labeling with self-attention.
\newblock In {\em Thirty-Second AAAI Conference on Artificial Intelligence},
  2018.

\bibitem{thai2018embedded}
Dung Thai, Sree~Harsha Ramesh, Shikhar Murty, Luke Vilnis, and Andrew McCallum.
\newblock Embedded-state latent conditional random fields for sequence
  labeling.
\newblock {\em arXiv preprint arXiv:1809.10835}, 2018.

\bibitem{tran2017named}
Quan Tran, Andrew MacKinlay, and Antonio~Jimeno Yepes.
\newblock Named entity recognition with stack residual lstm and trainable bias
  decoding.
\newblock {\em arXiv preprint arXiv:1706.07598}, 2017.

\bibitem{vaswani2016supertagging}
Ashish Vaswani, Yonatan Bisk, Kenji Sagae, and Ryan Musa.
\newblock Supertagging with lstms.
\newblock In {\em Proceedings of the 2016 Conference of the North American
  Chapter of the Association for Computational Linguistics: Human Language
  Technologies}, pages 232--237, 2016.

\bibitem{Vaswani2017Attention}
Ashish Vaswani, Noam Shazeer, Niki Parmar, Jakob Uszkoreit, Llion Jones,
  Aidan~N Gomez, {\L}ukasz Kaiser, and Illia Polosukhin.
\newblock Attention is all you need.
\newblock In {\em NIPS}, pages 5998--6008, 2017.

\bibitem{vinyals2015pointer}
Oriol Vinyals, Meire Fortunato, and Navdeep Jaitly.
\newblock Pointer networks.
\newblock In {\em Advances in Neural Information Processing Systems}, pages
  2692--2700, 2015.

\bibitem{wang2017named}
Chunqi Wang, Wei Chen, and Bo~Xu.
\newblock Named entity recognition with gated convolutional neural networks.
\newblock In {\em Chinese Computational Linguistics and Natural Language
  Processing Based on Naturally Annotated Big Data}, pages 110--121. Springer,
  2017.

\bibitem{wei2020position}
Wei Wei, Zanbo Wang, Xianling Mao, Guangyou Zhou, Pan Zhou, and Sheng Jiang.
\newblock Position-aware self-attention based neural sequence labeling.
\newblock {\em Pattern Recognition}, page 107636, 2020.

\bibitem{wu2018evaluating}
Minghao Wu, Fei Liu, and Trevor Cohn.
\newblock Evaluating the utility of hand-crafted features in sequence
  labelling.
\newblock {\em arXiv preprint arXiv:1808.09075}, 2018.

\bibitem{xin2018learning}
Yingwei Xin, Ethan Hart, Vibhuti Mahajan, and Jean-David Ruvini.
\newblock Learning better internal structure of words for sequence labeling.
\newblock {\em EMNLP}, 2018.

\bibitem{yadav2018survey}
Vikas Yadav and Steven Bethard.
\newblock A survey on recent advances in named entity recognition from deep
  learning models.
\newblock In {\em Proceedings of the 27th International Conference on
  Computational Linguistics}, pages 2145--2158, 2018.

\bibitem{yan2019tener}
Hang Yan, Bocao Deng, Xiaonan Li, and Xipeng Qiu.
\newblock Tener: Adapting transformer encoder for name entity recognition.
\newblock {\em arXiv preprint arXiv:1911.04474}, 2019.

\bibitem{yang2016multi}
Zhilin Yang, Ruslan Salakhutdinov, and William Cohen.
\newblock Multi-task cross-lingual sequence tagging from scratch.
\newblock {\em arXiv preprint arXiv:1603.06270}, 2016.

\bibitem{Yang2016Transfer}
Zhilin Yang, Ruslan Salakhutdinov, and William~W Cohen.
\newblock Transfer learning for sequence tagging with hierarchical recurrent
  networks.
\newblock In {\em ICLR}, 2017.

\bibitem{Yasunaga2018Robust}
Michihiro Yasunaga, Jungo Kasai, and Dragomir Radev.
\newblock Robust multilingual part-of-speech tagging via adversarial training.
\newblock In {\em NAACL}, 2018.

\bibitem{ye2018hybrid}
Zhi-Xiu Ye and Zhen-Hua Ling.
\newblock Hybrid semi-markov crf for neural sequence labeling.
\newblock {\em ACL}, 2018.

\bibitem{yu2015multi}
Fisher Yu and Vladlen Koltun.
\newblock Multi-scale context aggregation by dilated convolutions.
\newblock {\em arXiv preprint arXiv:1511.07122}, 2015.

\bibitem{zhai2017neural}
Feifei Zhai, Saloni Potdar, Bing Xiang, and Bowen Zhou.
\newblock Neural models for sequence chunking.
\newblock In {\em Thirty-First AAAI Conference on Artificial Intelligence},
  2017.

\bibitem{zhang2017embracing}
Boliang Zhang, Di~Lu, Xiaoman Pan, Ying Lin, Halidanmu Abudukelimu, Heng Ji,
  and Kevin Knight.
\newblock Embracing non-traditional linguistic resources for low-resource
  language name tagging.
\newblock In {\em Proceedings of the Eighth International Joint Conference on
  Natural Language Processing (Volume 1: Long Papers)}, pages 362--372, 2017.

\bibitem{Zhang2017Does}
Yi~Zhang, Xu~Sun, Shuming Ma, Yang Yang, and Xuancheng Ren.
\newblock Does higher order lstm have better accuracy for segmenting and
  labeling sequence data?
\newblock In {\em COLING}, 2017.

\bibitem{zhang2018learning}
Yuan Zhang, Hongshen Chen, Yihong Zhao, Qun Liu, and Dawei Yin.
\newblock Learning tag dependencies for sequence tagging.
\newblock In {\em IJCAI}, 2018.

\bibitem{zhang2018sentence}
Yue Zhang, Qi~Liu, and Linfeng Song.
\newblock Sentence-state lstm for text representation.
\newblock {\em ACL}, 2018.

\bibitem{Zhao2019ANM}
Sendong Zhao, Ting Liu, Sicheng Zhao, and Fei Wang.
\newblock A neural multi-task learning framework to jointly model medical named
  entity recognition and normalization.
\newblock {\em CoRR}, abs/1812.06081, 2019.

\bibitem{zheng2017joint}
Suncong Zheng, Feng Wang, Hongyun Bao, Yuexing Hao, Peng Zhou, and Bo~Xu.
\newblock Joint extraction of entities and relations based on a novel tagging
  scheme.
\newblock {\em arXiv preprint arXiv:1706.05075}, 2017.

\bibitem{zhou2002named}
GuoDong Zhou and Jian Su.
\newblock Named entity recognition using an hmm-based chunk tagger.
\newblock In {\em proceedings of the 40th Annual Meeting on Association for
  Computational Linguistics}, pages 473--480. Association for Computational
  Linguistics, 2002.

\bibitem{zhou2015answer}
Xiaoqiang Zhou, Baotian Hu, Qingcai Chen, Buzhou Tang, and Xiaolong Wang.
\newblock Answer sequence learning with neural networks for answer selection in
  community question answering.
\newblock {\em arXiv preprint arXiv:1506.06490}, 2015.

\bibitem{zhu2019can}
Yuying Zhu, Guoxin Wang, and B{\"o}rje~F Karlsson.
\newblock Can-ner: Convolutional attention network for chinese named entity
  recognition.
\newblock {\em NAACL}, 2019.

\bibitem{zhuo2016segment}
Jingwei Zhuo, Yong Cao, Jun Zhu, Bo~Zhang, and Zaiqing Nie.
\newblock Segment-level sequence modeling using gated recursive semi-markov
  conditional random fields.
\newblock In {\em Proceedings of the 54th Annual Meeting of the Association for
  Computational Linguistics (Volume 1: Long Papers)}, volume~1, pages
  1413--1423, 2016.

\end{thebibliography}
\bibliographystyle{plain}

\end{document}